\documentclass[12pt]{article}

\usepackage{epsfig}
\usepackage{color}

\usepackage[english]{babel}

\usepackage{fancyhdr}
\usepackage[normal]{subfigure}
\usepackage{amsmath}
\usepackage{amsfonts}
\usepackage{amssymb}
\usepackage{amscd}
\usepackage{latexsym}
\usepackage{graphicx}
\usepackage{cite}

\usepackage{a4wide}

\newcommand{\bd}{\begin{definition}}
	\newcommand{\ed}{\end{definition}}
\newcommand{\bt}{\begin{theorem}}
	\newcommand{\et}{\end{theorem}}
\newcommand{\bi}{\begin{itemize}}
	\newcommand{\ei}{\end{itemize}}
\newcommand{\ben}{\begin{enumerate}}
	\newcommand{\een}{\end{enumerate}}
\newcommand{\beq}{\begin{equation}}
	\newcommand{\eeq}{\end{equation}}

\newcommand{\re}{\mbox{$ \mathbb{R}  $}}

\newcommand{\igot}{\mbox{$\mathfrak{I}$}}
\newcommand{\sgn}{\mbox{${\rm sign}$}}

\newtheorem{definition}{Def.}[section]
\newtheorem{theorem}{Theorem}[section]
\newtheorem{lemma}{Lemma}[section]

\newtheorem{proposition}{Proposition}[section]

\def \proof{\noindent{\it Proof}.  \ignorespaces}
\def \qed{ \hfill $\Box$ \\}

\begin{document}
	
	\title{A Perceptually Inspired Variational Framework for Color Enhancement}
	\author{Rodrigo Palma-Amestoy, Edoardo Provenzi, \\ Marcelo Bertalm\'io and Vicent Caselles
		\thanks{Rodrigo Palma-Amestoy is with Department of
			Electrical Engineering, Universidad de Chile, Avenida Tupper 2007, Casilla 412-3, Santiago, Chile. E-mail: ropalma@ing.uchile.cl. Tel. : +56 2 978 4207. Fax. +56 2 695 3881. Edoardo Provenzi is with Dipartimento di Tecnologia dell'Informazione, Universit\`a di Milano, Via Bramante 65, 26013, Crema (CR),
			Italy. E-mail: provenzi@dti.unimi.it. Tel. +39 0373 898089. Fax.
			+39 0373 898010. Marcelo Bertalm\'{i}o and Vicent Caselles are with Departament de Tecnologia, Universitat Pompeu Fabra, Pg. de Circumvallacio 8, 08003 Barcelona, Spain. E-mail: marcelo.bertalmio@upf.edu, Tel. +34 93 542 25 69, Fax. +34  93 542 25 17; E-mail: vicent.caselles@upf.edu, Tel. +34  93 542 24 21, Fax. +34  93 542 25 17.} }
	
	\maketitle

	\begin{abstract}
		
		Basic phenomenology of human color vision has been widely taken as an inspiration to devise
		explicit color correction algorithms. The behavior of these models in terms of
		significative image features (such as contrast and dispersion)
		can be difficult to characterize. To cope
		with this, we propose to use a variational formulation of color contrast enhancement that is inspired
		by the basic phenomenology of color perception.
		In particular, we devise a set of basic requirements to be fulfilled by an energy
		to be considered as `perceptually inspired', showing that there is an explicit class
		of functionals satisfying all of them. We single out
		three explicit functionals that we consider of basic interest,
		showing similarities and differences with existing models.
		The minima of such functionals is computed using a gradient descent approach.
		We also present a general methodology to reduce the computational cost
		of the algorithms under analysis from ${\cal O}(N^2)$ to ${\cal O}(N\log N)$,
		being $N$ the number of input pixels.
		
	\end{abstract}
	
%
	\section{Introduction}
	
	Human vision is a process of great complexity that involves many
	processes such as shape and pattern recognition, motion
	analysis and color perception. An essential literature about
	neurophysiological aspects of human vision can be found in the
	books \cite{Gregory:97,Hubel:95,Palmer:99,Pratt:07,Wandell:95,Wyszecky:00,Zeki:93}.
	
	In this paper we will focus on color perception. This process
	begins with light capture by the three different types of cone
	pigments inside the retina. They have different absorption
	characteristics, with peak absorptions in the red, green and blue
	regions of the optical spectrum. The existence of three types of
	cones provides a physiological basis for the trichromatic theory
	of color vision. When a light stimulus activates a cone, a
	photochemical transition occurs, producing a nerve impulse.
	Experimental evidence \cite{Zeki:93} show that the propagation of
	nerve impulses reach the brain, which analyzes and interprets the
	information carried by the impulses. The part of the brain devoted
	to color perception is the so-called V4 zone of the occipital
	cortex. Neither retina photochemistry nor nerve impulses
	propagation are well understood, hence a \emph{deterministic}
	characterization of the visual process is still unavailable. For this
	reason, the majority of color perception models follow a
	\emph{descriptive} approach, trying to simulate macroscopic
	characteristics of color vision, rather than reproduce
	neurophysiological activity.
	
	The first modern contribution in such a field has been provided by
	the famous Retinex theory of Land and McCann \cite{Land:71}. They
	started devising a series of innovative experiments about color
	perception using unrealistic pictures called `Mondrians'
	\cite{Land:77,Land:83}. The abstract character of the images
	minimized the action of high level faculties of the Human Visual
	System (HVS) such as memory learning, judgement and recognition,
	thus making possible the reproducibility of the experiments
	\cite{Zeki:98}. They then found a formal computation able
	to reproduce experimental data, providing the first Retinex model
	of color vision \cite{Land:71}. After that, a whole category of
	perceptually inspired computational algorithms has been developed
	with different aims: reproduction of color sensation, color
	stabilization, enhancement of color images, etc. \cite{McCannat40:04}.
	A common characteristic of these works is that HVS features are taken as inspiration to devise the
	explicit equations of perceptually based algorithms.
	
	In this paper, rather than dealing directly with explicit algorithms, we are interested in the analysis of the interplay between color perception and variational image processing. The reason is that variational techniques have proven to be a powerful instrument to both \emph{understanding} and \emph{solving} image processing problems. Variational principles are based on energy functionals, whose selection depends on the particular problem under examination. The image for which the functional reaches a minimum is the solution to the particular image processing task at hand.
	Since our aim is performing a perceptually inspired color correction, we are interested in the analysis of energy functionals complying with the basic features of color perception. So, we will provide a translation of basic macroscopic \emph{phenomenological characteristics} of color vision into mathematical requirements to be fulfilled by an energy functional in order to be considered \emph{perceptually inspired}. Then, the minimization of any such functional starting from a given input image will produce an output image which will be a color-corrected version of the original, and this correction process will have followed (some) basic behavior of the HVS.
	
	We will show that it is possible to identify a class of
	functionals that comply with our set of basic perceptually based assumptions. Remarkably, this class
	happens to be the unique class that fulfills the logarithmic
	response to light stimuli expressed by Weber-Fechner's law. The Euler-Lagrange
	equations corresponding to the minimization of
	these functionals can be used to implement computational (gradient descent) algorithms. The
	fact that these equations come from a variational method permits to understand
	the algorithm's behavior in terms of contrast enhancement and control of dispersion (departure from original gray levels and average value.)
	We will present a theoretical and qualitative analysis of three
	particular examples of perceptually inspired energy functionals
	that we consider of basic interest. This analysis will put in evidence similarities and differences
	with Retinex and a more recent perceptual color correction
	algorithm called ACE \cite{Rizzi:03}. Furthermore, we will discuss a general procedure to reduce the
	computational cost of the algorithms derived by the three selected
	energy functionals from ${\cal O}(N^2)$ to ${\cal O}(N\log N)$,
	being $N$ the number of pixels in the input image.
	
	Let us finally describe the plan of the paper. In Section II we recall the basic HVS phenomenological properties that will lead us to formulate, in Section III, our basic set of assumptions that must be satisfied by an energy functional to be considered perceptually-inspired. Section IV contains three explicit examples of perceptual energies, the mathematical analysis of their variations is given in Section V. In Section VI we present a qualitative comparison between the algorithms corresponding to these energies and already existing perceptually-inspired algorithms. In Section VII we expose a general strategy to reduce the computational algorithm complexity. Section VIII contains the tests we have performed. We end with conclusions and open problems. All proofs have been put in the appendices.

	\section{Basic HVS phenomenological properties}
	
	The phenomenological properties of human color perception that we
	consider as basics for our axiomatic construction are three: color
	constancy, local contrast enhancement and visual adaptation. For
	the sake of clarity, we discuss them in different subsections.

	\subsection{Human color constancy}
	
	It is well known that the HVS has strong `normalization'
	properties, i.e. humans can perceive colors of a scene almost
	independently of the spectral electromagnetic composition of a
	uniform illuminant, usually called color cast
	\cite{Land:71,Land:77}. This peculiar ability is known as
	\emph{human color constancy} and it is not perfect
	\cite{West:79,Funt:98}. In fact it depends on several
	factors, such as cast strength, or amount of detail in the
	scene. In particular, if the variety of visual information is
	poor, the HVS can be deceived. In fact, with a suitable selection
	of the illuminant, two perfectly uniform surfaces with different
	reflectance can be perceived as having the same color.
	However, as proved by Land's experiments \cite{Land:71}, this cannot happen if
	the surface is not uniform, showing that human color perception
	strongly depends on the context, i.e. on the \emph{relative}
	luminance coming from different places of an observed scene,
	rather than on the \emph{absolute} luminance value of every single
	point. Nonetheless, being color constancy not complete, also
	absolute punctual luminance information plays a role in the entire
	color perception process and cannot be completely discarded.
	
\begin{figure}[!h]
		\center
		\includegraphics[width=2in]{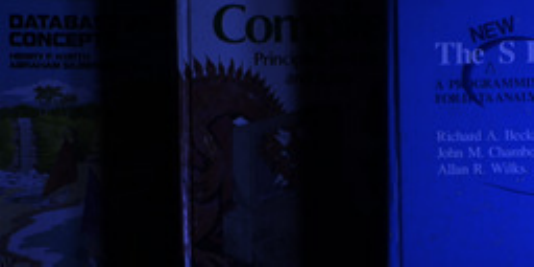}\,
		\includegraphics[width=3in]{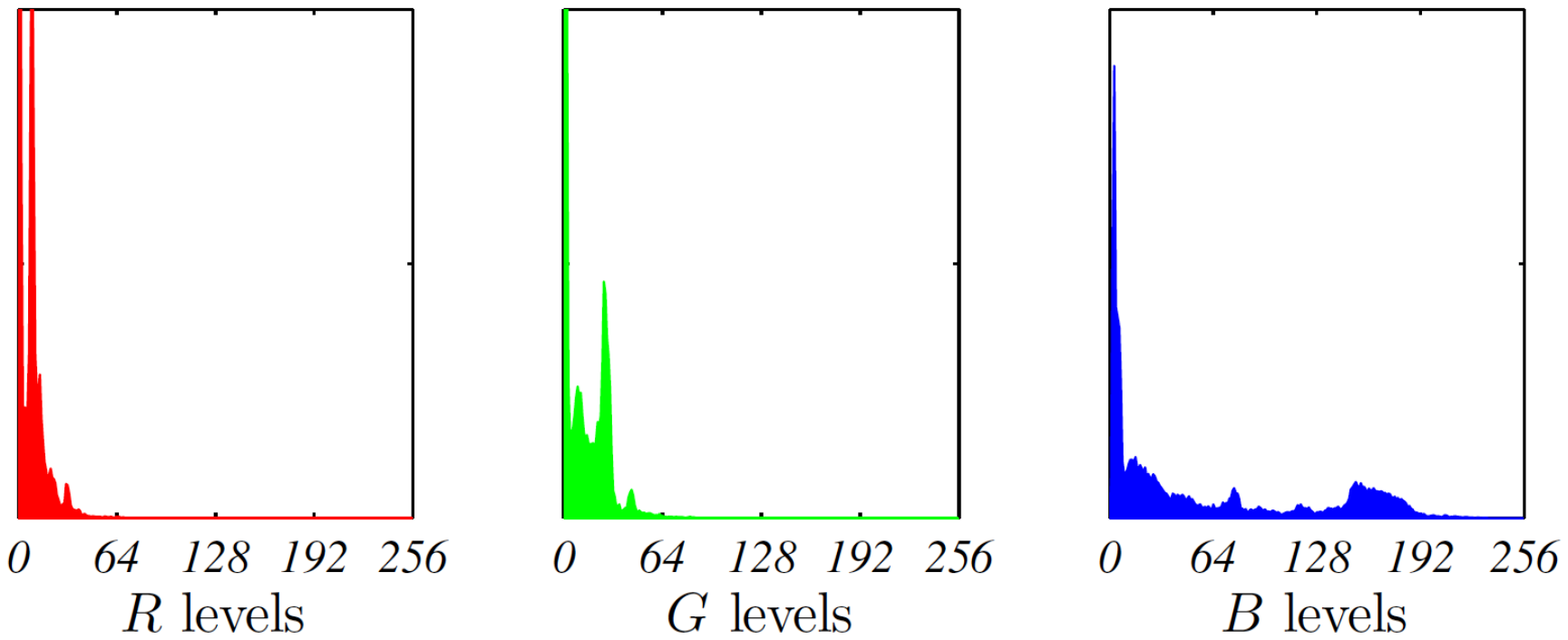}\,
		\caption{A picture with blue cast and its global R,G,B histograms.} \label{fig:book}
	\end{figure}

	The majority of color models try to remove the color cast due to illuminant looking for the invariant component of the light signal: the physical reflectance of objects (see \cite{Ebner:07} for a recent overview of color constancy algorithms). However it is well known that the separation between illuminant and reflectance is an ill-posed problem, unless one imposes further constraints which are not verified by all images \cite{Hurlbert:86}.
	We will see later that we intend to remove color cast using another approach based on contrast enhancement. To understand why we relate color cast and contrast enhancement, we notice that an image with color cast is always
	characterized by one chromatic channel with remarkably different
	standard deviation with respect to the others (not to be confused with an image with dominant color, e.g. the close up of a leaf, in which is the average value of a channel that prevails). To have
	a quantitative example let us consider, for instance, the picture
	with strong blue cast shown in fig. \ref{fig:book} with the
	relative R,G,B histograms. The average values of R,G,B
	intensities are, respectively, $\mu_R=8.67$, $\mu_G=14.79$,
	$\mu_B=60.18$, while the standard deviations are $\sigma_R =
	17.59$, $\sigma_G =27.03$, $\sigma_B =130.56$.
	
	An image without color cast does not show such a of difference between standard deviations in the three chromatic channels. Being the standard deviation a measure of average contrast, it is clear that contrast enhancement can help decreasing the difference between $\sigma_R$, $\sigma_G$ and $\sigma_B$, by spreading the intensity values of all separate chromatic channels, thus reducing color cast.

	\subsection{Local contrast enhancement}
	
	The HVS provides a local contrast enhancement, in the sense that
	the perception of details is spatially dependent. Well
	known phenomena exhibiting local contrast enhancement are Mach
	bands and simultaneous contrast \cite{Gonzales:02}. Local HVS features related to
	color perception can be appreciated also in more complex scenes
	and go under the general name of `chromatic induction'
	\cite{Creutzfeld:87,Creutzfeld:90,Hurvich:90,Zaidi:99}. Experimental
	evidences show that the strength of chromatic induction between
	two different areas of a scene decreases monotonically with their
	Euclidean distance, even though a precise analytical description
	is not yet available.

	\subsection{Visual adaptation}
	
	The range of light intensity levels to which the HVS can adapt is
	impressive, on the order of $10^{10}$, from the scotopic threshold
	to the glare limit \cite{Gonzales:02}. However, the HVS system cannot
	operate over such a range simultaneously, rather it adapts to each
	scene to prevent saturation which would depress contrast
	sensitivity \cite{Shapley:84}.
	
	During this adaptation, the HVS \emph{shifts} its sensitivity to
	light stimuli in order to present only modulations around the
	average intensity level of a scene \cite{Shapley:84}. This provides a
	phenomenological motivation for the so-called `gray-world' (GW)
	assumption, which says that the average color in a scene is gray
	\cite{Buchsbaum:80}.

	\section{The basic set of assumptions}\label{sect:assumptions}

	We now want to use the phenomenological characteristics just
	mentioned to write a set of basic assumptions to be satisfied by a perceptually inspired energy
	functional. Let us fix the notation that will be used throughout
	the paper. For notational simplicity, and also motivated by practical applications,
	we shall work with discrete images, even if many of the formulae we will consider could be written in an analogous context
	by replacing sums by integrals.
	
	Given a discrete RGB image, we denote by $\mathfrak{I}$
	$=\{1,\ldots, W\} \times \{1,\ldots, H\} \subset {\mathbb{Z}}^2$ its spatial domain, $W, H \geq 1$ being integers;
	$x = (x_1,x_2)$ and $y = (y_1,y_2)$ denote the coordinates of two
	arbitrary pixels in $\mathfrak{I}$. We will always consider a
	normalized dynamic range in $[0,1]$, so that a color image function is denoted by $\vec{I}:{\mathfrak{I}}
	\to [0,1]^3$, $\vec{I}(x)=(I_R(x),I_G(x),I_B(x))$, where $I_k(x)$
	is the intensity level of the pixel $x\in \mathfrak I$ in the
	chromatic channel $k \in \{R,G,B\}$. All computations will be
	performed on the scalar components of the image, thus treating
	independently each chromatic channel. As will be later explained with more detail,
	this choice permits us to properly treat
	images with color cast (see \cite{Hubel:95} chapter 8 `Color Vision'
	for more details about independent chromatic channel computation
	and color cast).
	
	For later convenience, we propose to use the following convention.
	Given any image $I: {\mathfrak{I}} \to [0,1]$ we extend it as an even
	function with respect to the two variables in the domain $\{-W +
	1,\ldots, W\} \times \{-H + 1,\ldots, H\}$ (which we still indicate with $\mathfrak{I}$ for simplicity), i.e. we replicate the image specularly in all directions, and then by periodicity
	to ${\mathbb{Z}}\times {\mathbb{Z}}$ with fundamental period $\mathfrak{I}$.
	For simplicity, we denote the extended image again by $I$. With this, we may consider
	the domain of $I$ as the periodic sampling lattice, that is
	${\mathbb{T}}_d : = ({\mathbb{Z}}\times {\mathbb{Z}})/$
	$(2W\mathbb{Z}$ $\times 2H\mathbb{Z})$. This notation means
	that we identify any pair of points $x=(x_1,x_2)$ and
	$y=(y_1,y_2)$ in ${\mathbb{Z}}\times {\mathbb{Z}}$ if $x_1-y_1 \in
	2W\mathbb{Z}$ and $x_2-y_2 \in 2H\mathbb{Z}$. We denote this
	equivalence relation by $\equiv$. The distance between any two
	points $x,y \in \mathbb{T}_d$, denoted by $\Vert x - y
	\Vert_{{\mathbb{T}}_d} $, is computed as $\min\{ \vert \tilde{x}
	-\tilde{y}\vert: \tilde{x}\equiv x,\, \tilde{y}\equiv y\}$, where
	$\vert v \vert = \sqrt{v_1^2+v_2^2}$, $v = (v_1,v_2)$. From now on,
	we shall assume that our images have these symmetry and
	are defined on the extended domain $\mathfrak{I}$.
	
	
	\subsection{The first assumption: General structure of the energy functional}
	
	In order to write the general structure of a perceptually inspired
	energy functional, we start noticing that human color perception is characterized by \emph{both local and global features}:
	contrast enhancement has a local nature, i.e. \emph{spatially
		variant}, while visual adaptation and attachment to original data (implied by the failure of color constancy) have a global nature, i.e. \emph{spatially invariant}, in the sense that they do not depend on the intensity distribution in the neighborhood.
	
	These basic considerations imply that a perceptually inspired energy functional should contain two terms: one spatially-dependent term whose minimization leads to a local contrast enhancement and one global term whose minimization leads to a control of the departure from both original punctual values and the middle gray, which, in our normalized dynamic range, is $1/2$.
	
	Let us first describe the general form of the contrast enhancement terms
	that we will consider in this paper. For that we need a contrast measure $\overline{c}(a,b)$ between two gray levels
	$a,b > 0$ (to avoid some singular cases, we shall assume that intensity image values are always positive).
	We require the contrast function
	$\overline{c}:(0,+\infty) \times (0,+\infty) \to \mathbb{R}$
	to be continuous, symmetric in $(a,b)$, i.e.
	$\overline{c}(a,b) = \overline{c}(b,a)$, increasing when $\min(a,b)$ decreases or $\max(a,b)$ increases.
	Basic examples of contrast measures are $\overline{c} = \vert a - b\vert \equiv \max(a,b) - \min(a,b)$ or $\overline{c}(a,b) = \frac{\max(a,b)}{\min(a,b)}$.
	
	Since our purpose is to enhance contrast by \emph{minimizing} an energy, we define
	an \emph{inverse contrast function} $c(a,b)$, still continuous and symmetric in $(a,b)$, but
	decreasing when $\min(a,b)$ decreases or $\max(a,b)$ increases. Notice that,
	if $\overline{c}(a,b)$ is a contrast measure, then $c(a,b) = - \overline{c}(a,b)$
	or $c(a,b) = 1/\overline{c}(a,b)$ is an inverse contrast measure, so that basic examples of inverse contrast are: $c(a,b)=\min(a,b) - \max(a,b)$ or
	$c(a,b)=\frac{\min(a,b)}{\max(a,b)}$.


Let us now introduce a weighting function to localize the contrast computation. Let $w: \mathfrak{I} \times \mathfrak{I} \to \mathbb{R}^+$ be a positive symmetric kernel, i.e. such that
$w(x,y) = w(y,x) > 0$, for all $x,y\in \igot$, that measures the mutual influence between the pixels $x,y$. The symmetry requirement is motivated by the fact that the mutual chromatic induction is independent on the order of the two pixel considered. Usually, we assume that
$w(x,y)$ is a function of the Euclidean distance $\Vert x-y\Vert_{\mathfrak{I}}$ between the two points. We shall assume that the kernel is normalized, i.e. that
\begin{equation}\label{eq:norma}
	\sum_{y\in \mathfrak{I}} w(x,y)  = 1 \qquad \forall x\in \mathfrak{I}.
\end{equation}

Given an inverse contrast function $c(a,b)$ and a positive symmetric kernel $w(x,y)$, we define a contrast energy term
by
\begin{equation}\label{def:contrastterm}
	C_w(I)=\sum_{x\in \mathfrak I} \sum_{y\in \mathfrak I} w(x,y) \,
	c(I(x),I(y)) \, .
\end{equation}

Thanks to the symmetry assumption, we may write $c(a,b) = \tilde{c} (\min (a,b), \max (a,b))$
for some function $\tilde{c}$ (indeed well defined by this identity). Notice that $\tilde{c}$ is non-decreasing in
the first argument and non-increasing in the second one. The symmetry hypothesis is not restrictive, in fact, if the inverse contrast measure $c(a,b)$ were not symmetric, we could write it as the sum $c(a,b) = c_s(a,b) + c_{as}(a,b)$ where $c_s(a,b)$ and $c_{as}(a,b)$ are symmetric and anti-symmetric respectively. Since the sum $\sum_{ (x,y) \in \mathfrak I^2} w(x,y) \, c_{as}(I(x),I(y)) \,  = 0$, then the only remaining term is
$\sum_{(x,y)\in \mathfrak I^2} w(x,y) \, c_{s}(I(x),I(y)) \,$, hence we may assume that $c(a,b)$ is symmetric in $(a,b)$.

Let us now consider the term that should control the dispersion. As suggested previously, we intend it as an attachment term to
the initial given image $I_0$ and to the average illumination value, which we assume
to be $1/2$. Thus, we define two dispersion functions:
$d_1(I(x),I_0(x))$ to measure the separation between $I(x)$ and $I_0(x)$, and
$d_2(I(x),\frac{1}{2})$ which measures the
separation from the value $1/2$. Both $d_1$ and $d_2$ are continuous functions $d_{1,2}:{\mathbb{R}}^2 \to {\mathbb{R}}^+$
such that $d_{1,2}(a,a) = 0$ for any $a \in \mathbb{R}$, and $d_{1,2}(a,b) > 0$ if $a \neq b$.
We write $d_{I_0,\frac{1}{2}}(I(x)) = d_1(I(x),I_0(x)) + d_2(I(x),\frac{1}{2})$, and
the dispersion energy term as
\begin{equation}\label{def:dispersionterm}
	D(I)=\sum_{x\in \mathfrak I}  d_{I_0,\frac{1}{2}}(I(x)) \, .
\end{equation}


We can now formulate our first assumption.

\medskip
\medskip

\noindent {\em Assumption 1. The general structure of a perceptually inspired
	color correction energy functional is
	\beq\label{def:energyCD} E_{w}(I) = D(I) +
	C_w(I), \eeq
	where $C_w(I)$ and $D(I)$ are the contrast and dispersion terms defined in
	(\ref{def:contrastterm}) and (\ref{def:dispersionterm}), respectively.
	The minimization of $D$ must provide a
	control of the dispersion around 1/2 and around the original
	intensity values. The minimization of $C_w$ must provide a local
	contrast enhancement.}

\medskip
\medskip

We notice that, even if we have derived the above structure of a perceptual energy functional starting form phenomenological evidence, it agrees with the general structure proposed in \cite{Usui:97} on the basis of neurophysiological evidences.

\subsection{The second assumption: Properties of the contrast function}


In order to find out which properties the contrast term should
satisfy, let us observe that an overall change in intensity, measured by the generic quantity $\lambda >0$, does not affect the visual sensation. This requires the contrast function $c$ to be homogeneous, recalling that $c$ is homogeneous of degree $n\in \mathbb Z$ if
\beq \label{eq:homogeneous}
c(\lambda a,\lambda b) = \lambda^n \, c(a,b) \quad \forall \lambda, a,b\in (0,+\infty),
\eeq
where $a$ and $b$ are synthetic representations of $I(x)$ and $I(y)$.
Of course, if $n=0$, $c$ automatically disregards the presence of $\lambda$, but we can say more: since $\lambda$ can take any positive value, if we set $\lambda = 1/b$, we may write equation (\ref{eq:homogeneous}) as:
\beq
c(a,b) = b^n \, c\left(\frac{a}{b},1\right) \quad \forall a,b\in (0,+\infty),
\eeq
so, when $n=0$, $b^n = 1$ and $c$ results as a function of the ratio $a/b$ which intrinsically disregards overall changes in light intensity. If $n>0$, then $\lambda$ has a global influence and could be removed performing a
suitable normalization (for instance, dividing by the $n$-th power of the highest intensity level). We can formalize these considerations in our second assumption.

\medskip
\medskip
\noindent {\em Assumption 2. We assume that the inverse contrast function $c(a,b)$ is homogeneous.}
\medskip
\medskip
\noindent

Thanks to the arguments presented so far, we have that inverse contrast functions which are
homogeneous of degree $n=0$ are those that can be written as a monotone non-decreasing function of
$\frac{\min(I(x),I(y))}{\max(I(x),I(y))}$. It would be interesting to know if,
given an inverse contrast function  $c$, there are increasing functions $g,h:(0,+\infty)\to(0,+\infty)$
such that $g(c(h(a),h(b))) = \frac{a}{b}$, for any $0 < a < b$. Probably some other assumptions on $c$ are required to
obtain such a characterization, which means that, modulo a calibration of intensity values, $\frac{\min(I(x),I(y))}{\max(I(x),I(y))}$ is the essential inverse contrast function. In the next subsection we provide further support to restrict ourselves to such class of contrast functions.

\subsection{Contrast functions satisfying the Weber-Fechner's law}

Weber-Fechner's law describes a common behavior of human senses.
It has been first discovered by Weber and then formalized by
Fechner. In general, it describes a non-linear response of a human
sense to variations of external stimuli: the same variation is
perceived in a weaker way as the strength of the external stimulus
increases. The experiment for human vision is performed in
controlled conditions \cite{Gonzales:02}: on a uniform background of
intensity $I_0$, it is superimposed a brief pulse of light of
intensity $I_1 > I_0$. Weber-Fechner's law states that the so-called
\emph{Weber-Fechner ratio} ${\cal R}_{WF}\equiv
\frac{I_1-I_0}{I_0}$, i.e. the ratio between the variation
$\Delta I \equiv I_1-I_0$ and the background $I_0$, remains
constant. Even though Weber-Fechner's law is not perfect
\cite{Pratt:07}, the intensity range over which it is in good
agreement with experience (called `Weber-Fechner's domain') is
still comparable to the dynamic range of most electronic imaging
systems \cite{Pratt:07}.

Since ${\cal R}_{WF} = I_1/I_0 -1 $, Weber-Fechner's law is saying that
the perceived contrast is a function of $I_1/I_0$. This reason motivates us to say that
$c(a,b)$ is a {\em generalized Weber-Fechner contrast function} if $c$ is an \emph{inverse contrast}
function which can be written as a non-decreasing function of $\min(a,b)/\max(a,b)$. Hence, we can particularize
assumption 2 as follows.

\medskip
\medskip
{\em Assumption 2'. We assume that $c$ is a generalized Weber-Fechner contrast function.}

\medskip
\medskip

\section{Explicit energy functionals complying with our set of assumptions}\label{sect:energies}

The purpose of this section is to show explicit examples of energy
functionals complying with our set of assumptions. We first concentrate on the contrast terms, then we
discuss the dispersion terms and, finally, we analyze the
Euler-Lagrange equations of the complete energy functionals. Here
we mainly concentrate on mathematical features, reserving
qualitative interpretations and comparisons for the following
section.

\subsection{Three basic inverse contrast terms}


As we have explained in Section \ref{sect:assumptions} we shall concentrate on
generalized Weber-Fechner contrast terms. The following three are of particular interest:
\beq \label{eq:id} C^{\, \rm id}_w(I):=\frac{1}{4}
\sum_{x\in \mathfrak I} \sum_{y\in \mathfrak I} w(x,y) \,
\frac{\min(I(x),I(y))}{\max(I(x),I(y))} \, ,
\eeq
\beq
\label{eq:log} C^{\, \log}_w(I):=\frac{1}{4} \sum_{x\in \mathfrak I} \sum_{y\in \mathfrak I} w(x,y) \, \log \left(\frac{\min(I(x),I(y))}{\max(I(x),I(y))}\right) \, ,
\eeq
\beq \label{eq:m} C^{\, -{\cal M}}_w(I):=-\frac{1}{4}
\sum_{x\in \mathfrak I} \sum_{y\in \mathfrak I} w(x,y) \, {\cal
	M}\left(\frac{\min(I(x),I(y))}{\max(I(x),I(y))}\right) \, ,
\eeq
where
\beq\label{eq:defmichelson}{\cal
	M}\left(\frac{\min(I(x),I(y))}{\max(I(x),I(y))}\right):=
\frac{1-\frac{\min(I(x),I(y))}{\max(I(x),I(y))}}{1+\frac{\min(I(x),I(y))}{\max(I(x),I(y))}}
\equiv \frac{|I(x)-I(y)|}{I(x)+I(y)},
\eeq
is the well known Michelson's definition of contrast \cite{Michelson:1927}. The upper
symbol in the above definitions of $C_w$ simply specifies the monotone function applied on the basic contrast variable $t: = \frac{\min(I(x),I(y))}{\max(I(x),I(y))}$, i.e.
identity $\mathrm{id}(t)=t$, logarithm $\log t$, and minus the Michelson's contrast function $-{\cal M}(t)= - \frac{1-t}{1+t}$,
respectively. Indeed, these three functions are strictly increasing
(this is obvious for the first two, for the third observe that $- {\cal M}^\prime(t)= 2/(1+t)^2 > 0$ for all $t$).
Notice that the function $t = \min(I(x),I(y))/\max(I(x),I(y))$ is minimized when
$\min(I(x),I(y))$ takes the smallest possible value and $\max(I(x),I(y))$ takes the largest
possible one, which corresponds to a contrast stretching. Thus, minimizing an increasing function of the
variable $t$, will produce a contrast enhancement. To refer to any one of them we use the notation $C_w^f(I)$,
where $f={\rm id}, {\rm log}, -{\cal M}$.


Let us briefly comment on the reasons for considering those contrast terms among others
in the Weber-Fechner class.
The first one is the simplest, and it is useful to understand
the basic contrast enhancement
properties of our energy functionals. The second has been chosen
because of the logarithmic property to transform ratios in
differences, so that the basic contrast variable becomes
$\min(\tilde I(x),\tilde I(y)) - \max(\tilde I(x),\tilde I(y)) = - \vert \tilde I(x) - \tilde I(y) \vert $,
being $\tilde I \equiv \log I$. Finally, the third one has been chosen
because it is given in terms of the integral of
Michelson's contrast.


%

\subsection{Entropic dispersion term}


The main features of the dispersion term have to be its attachment to
the initial given image $I_0$ and to the average illumination value, which we assume
to be $1/2$. In principle, to measure the dispersion of $I$ with respect to $I_0$ or $1/2$, any distance function can be used.
The simplest example would be a quadratic distance
\begin{equation}\label{qd1}
	D^q_{\alpha,\beta}(I):=\frac{\alpha}{2}\sum_{x\in \mathfrak I}
	\left( I(x) - \frac{1}{2} \right)^2 +
	\frac{\beta}{2}\sum_{x\in \mathfrak I} \left( I(x) - I_0(x) \right)^2
	, \qquad \alpha,\beta > 0.
\end{equation}
Given that contrast terms are expressed as homogeneous functions
of degree 0, the variational derivatives are
homogeneous functions of degree -1 (see Appendix VIII-C). Since our axioms do not give
any precise indication about the analytical form of the dispersion
term that should be chosen, we search for functions able to
maintain coherence with this homogeneity.
A good candidate for this is
the \emph{entropic dispersion} term, i.e.
\beq D_{\alpha,\beta}^{\cal E}(I):= \alpha
\sum_{x\in \mathfrak I} \left(\frac{1}{2} \log \frac{1}{2I(x)} -
\left(\frac{1}{2}-I(x)\right)\right)  + \beta \sum_{x\in \mathfrak I}
\left(I_0(x) \log \frac{I_0(x)}{I(x)} -
\left(I_0(x)-I(x)\right)\right) ,\eeq
where $\alpha,\beta > 0$, which is based on the relative entropy distance \cite{Ambrosio:05} between $I$ and
$1/2$ (the first term) and between $I_0$ and $I$(the second term).
Notice that, if $a > 0$ and $f(s)= a \log \frac{a}{s} -
(a-s)$, $s \in (0,1]$, then $\frac{df}{ds}(s)=1-\frac{a}{s}$ and
$\frac{d^2f}{ds^2}(s)=\frac{a}{s^2}>0$, $\forall s$. So, $f(s)$ has a
global minimum in $s=a$.
In particular, this holds when $a= I_0(x)$ or $a=1/2$. Given the statistical
interpretation of entropy, we can say that \emph{minimizing
	$D_{\alpha,\beta}^{\cal E}(I)$ amounts to minimizing the disorder of intensity levels
	around 1/2 and around the original data $I_0(x)$}. Thus, $D_{\alpha,\beta}^{\cal E}(I)$ accomplishes the
required tasks of a dispersion term.

\begin{figure}[!h]
	\centering
	\includegraphics[width=3in]{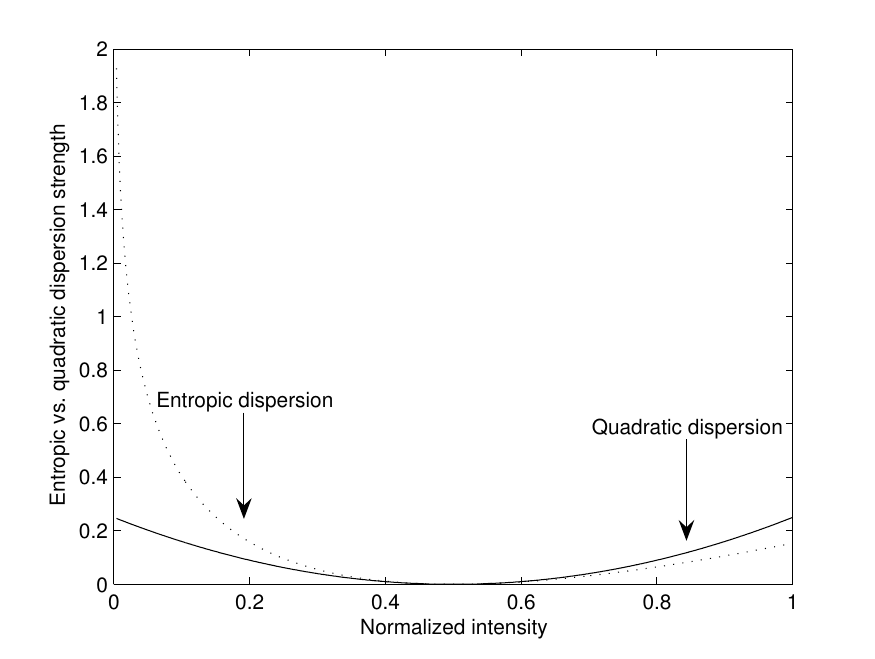}
	\caption{\emph{Dotted line}. Graphic of entropic dispersion function around 1/2.
		\emph{Solid line}. Graphic of quadratic dispersion function around 1/2.}\label{fig:dispersions}
\end{figure}

Notice that the integrand
functions of the entropic dispersion term are not symmetric with
respect to their minima, as instead happens for the ones of the
quadratic dispersion term
$D^q_{\alpha,\beta}(I)$. The difference can be appreciated in the graphics of
Fig.\ref{fig:dispersions} which refer, for simplicity, to the
dispersion around 1/2.

In the following section we show the explicit minimization of these energy functionals.


\section{Minimization of the energy functionals $E^{f}_{w,\alpha,\beta}(I)=D_{\alpha,\beta}^{\cal E}(I) + C_w^{f}(I)$}
\label{subsect:minEf}

Recall that the minimization of $E^{f}_{w,\alpha,\beta}(I)$, $f= {\rm id}, {\rm log}, - {\cal M}$
corresponds to
a trade-off between two opponent mechanisms: on one hand we have entropic
control of dispersion around 1/2 and around original
data, on the other hand we have local contrast enhancement.

The existence of a minimum in the discrete framework can be guaranteed for a quite general class
of energy functionals. Let us define, for $\rho > 0$, ${\cal F}_\rho : = \{I:{\mathfrak{I}} \to [0,1], \, I(x) \geq \rho \; \forall x\in \igot\}$. We minimize the energy $E$ in the class ${\cal F}_\rho$ to avoid singularities when we use $f={\rm log}$ or in the entropic dispersion term.

\medskip
\medskip

\begin{proposition}\label{prop:existence}
	Let $c:(0,1]\times (0,1] \to \mathbb{R}$ be a continuous function. Let
	$E(I) := D_{\alpha,\beta}^{\cal E}(I) + C_w(I)$
	where $C_w(I)$ is given in (\ref{def:contrastterm}). There is a minimum of $E(I)$
	in the class of functions ${\cal F}_\rho$.
	An analogous result holds if we use the quadratic dispersion term $D_{\alpha,\beta}^q(I)$.
\end{proposition}

\medskip
\medskip
The proof is given in Appendix VIII-A.
\medskip

Notice that if the energy $E$ is differentiable the minimum $I^*$ satisfies
$\delta E(I^*) = 0$. Before computing the Euler-Lagrange equations,
we notice that the contrast terms $C_w^{f}(I)$, $f= {\rm id}, {\rm log}, - {\cal M}$ are not convex
and the basic function $t: = \frac{\min(a,b)}{\max(a,b)}$ is not differentiable. In fact, we may write
$\min(a,b) = \frac{1}{2} (a+b- \vert a - b\vert)$,
$\max(a,b) = \frac{1}{2} (a+b+\vert a - b\vert)$, for any $a,b\in\mathbb{R}$.
The non-differentiability comes from the absolute value
$A(z) = \vert z\vert$, $z\in\mathbb{R}$.
Since our algorithm will use a gradient descent approach, we must regularize
the basic variable $t$.
We notice that $A^\prime(z) = 1$ if $z >0$,
$A^\prime(z) = -1$ if $z < 0$ and $A$ is not differentiable at $z=0$. But
all the values $s\in [-1,1]$ are \emph{subtangents} of $A(z)$ at $z=0$, that is,
$A(z) - A(0) \geq s(z-0)$ for any $z \in \mathbb{R}$. Thus we may write
$A^\prime (z) = {\rm sign}(z)$, where
\begin{eqnarray}\label{def:sign}
	{\rm sign}(z) = \left\{ \begin{array}{ll}
		1 & {\rm if}\, \,  z> 0 \\
		{ [-1,1]} & {\rm if}\, \,  z = 0 \\
		-1 & {\rm if}\, \,   z< 0
	\end{array} \right. .
\end{eqnarray}
We define ${\rm sign}_0(z)$ as in (\ref{def:sign}), but with the particular choice $0$ when $z=0$.

\medskip
\medskip

\begin{definition}\label{def:regulaAbsvalue}
	Given $\epsilon > 0$, we say that $A_\epsilon(z)$ is a `\emph{nice regularization}' of $A(z)$, if
	$A_\epsilon(z) \geq 0$ is convex, differentiable with continuous derivative,
	$A_\epsilon(0) = 0$, $A_\epsilon(-z)=A_\epsilon(z)$, and
	
	\noindent $(i)$ $A_\epsilon(z)\leq \vert z\vert$ for any $z\in\mathbb{R}$ and $A_\epsilon(z) = \vert z\vert + Q_{1,\epsilon}(z)$
	where $Q_{1,\epsilon}(z) \to 0$ as $\epsilon \to 0$, uniformly in $z\in [-1,1]$;
	
	\noindent $(ii)$ Let us denote $s_\epsilon(z) = A_\epsilon^\prime(z)$. Then
	$\vert s_\epsilon(z)\vert \leq 1$ for any $z\in [-1,1]$, $s_\epsilon(z) \to {\rm sign}_0(z)$  as $\epsilon \to 0$ for any $z\in \mathbb{R}$, and
	$ Q_{2,\epsilon}(z) : = A_\epsilon(z) - z s_\epsilon(z)\to 0$ as $\epsilon \to 0$, uniformly in $z\in [-1,1]$.
\end{definition}

\medskip
\medskip

We present two examples of nice regularization of $A(z)$.
\emph{Example} a): $A_\epsilon(z) = \sqrt{\epsilon^2+\vert z\vert^2}-\epsilon$, in this case $s_\epsilon(z) = \frac{z}{\sqrt{\epsilon^2+\vert z\vert^2}}$,
$Q_{1,\epsilon}(z) = O(\epsilon)$ and $Q_{2,\epsilon}(z) := A_\epsilon(z) - z s_\epsilon(z) = O(\epsilon)$.
\emph{Example} b):  $A_\epsilon(z) = z \frac{\arctan(z/\epsilon)}{\arctan (1/\epsilon)} -
\frac{\epsilon}{2\arctan (1/\epsilon)} {\rm log} (1+\frac{z^2}{\epsilon^2})$, in this case
$s_\epsilon(z) = \frac{\arctan(z/\epsilon)}{\arctan (1/\epsilon)}$,
$Q_{1,\epsilon}(z) = O(\epsilon \log \,  (1/\epsilon))$ and
$Q_{2,\epsilon}(z) = O(\epsilon \log \,  (1/\epsilon))$, uniformly in $z\in [-1,1]$.
We have denoted by $O(F(\epsilon))$ any expression satisfying $\vert O(F(\epsilon))\vert \leq C F(\epsilon)$
for some constant $C > 0$ and $\epsilon > 0$ small enough.
Observe that, in both cases $s_\epsilon(z) \to {\rm sign}_0(z)$  as $\epsilon \to 0$ for any $z\in \mathbb{R}$.
We give the details of the above statements in Appendix VIII-B.

\medskip Now let us assume that $A_\epsilon(z)$ is a nice regularization of $A(z)$. We set
\beq\label{def:maxminepsilon}
{\rm min}_\epsilon (a,b) = \frac{1}{2} (a+b - A_\epsilon(a-b)), \qquad
{\rm max}_\epsilon (a,b) = \frac{1}{2} (a+b + A_\epsilon(a-b)).
\eeq
We define the regularized version of the functionals as:
\beq \label{eq:ide} C^{\, \rm id}_{w,\epsilon}(I):=\frac{1}{4}
\sum_{x\in \mathfrak I} \sum_{y\in \mathfrak I} w(x,y) \,
\frac{{\rm min}_\epsilon(I(x),I(y))}{{\rm max}_\epsilon(I(x),I(y))} \, ;
\eeq
\beq
\label{eq:loge} C^{\, \log}_{w,\epsilon}(I):=\frac{1}{4} \sum_{x\in \mathfrak I} \sum_{y\in \mathfrak I} w(x,y) \, \log \left(
\frac{{\rm min}_\epsilon(I(x),I(y))}{{\rm max}_\epsilon(I(x),I(y))}\right) \, ;
\eeq
\beq \label{eq:me} C^{\, -{\cal M}}_{w,\epsilon}(I):=-\frac{1}{4}
\sum_{x\in \mathfrak I} \sum_{y\in \mathfrak I} w(x,y) \, \frac{A_\epsilon(I(x)-I(y))}{I(x)+I(y)}\, .
\eeq

\medskip

\begin{proposition}\label{prop:var_minmax}
	Assume that $A_\epsilon(z)$ is a nice regularization of $A(z)$.
	
	\medskip\noindent
	$(i)$ The first variation of $C^{\, \rm id}_{w,\epsilon}(I)$ is:
	\beq \begin{array}{ll}\delta C^{\,
			\rm id}_{w,\epsilon}(I) = - \frac{1}{2}\sum_{y\in \mathfrak I} w(x,y) \frac{I(y)}{{\rm max}_\epsilon(I(x),I(y))^2}s_\epsilon(I(x)-I(y))
		+ S_\epsilon  \\
		\\
		\qquad \qquad  = - \frac{1}{2}\sum_{y\in \mathfrak I} w(x,y) \frac{I(y)}{{\rm max}(I(x),I(y))^2}s_\epsilon(I(x)-I(y))
		+ S'_\epsilon,
	\end{array}
	\eeq
	where $S_\epsilon,S'_\epsilon= O(Q_{1,\epsilon}(I(x)-I(y))+ Q_{2,\epsilon}(I(x)-I(y)))$; notice the difference
	between $\max$ and $\max_\epsilon$ in the previous formula;
	
	\noindent \medskip \medskip $(ii)$ The first variation of $C^{\, \log}_{w,\epsilon}(I)$ is:
	\beq \delta C^{\,
		\log}_{w,\epsilon}(I) = - \frac{1}{2}\sum_{y\in \mathfrak I} w(x,y) \frac{1}{I(x)}s_\epsilon(I(x)-I(y))
	+ S_\epsilon,
	\eeq
	where $S_\epsilon= O(Q_{1,\epsilon}(I(x)-I(y))+ Q_{2,\epsilon}(I(x)-I(y)))$;
	
	\noindent \medskip \medskip $(iii)$ The first variation of $C^{-{ \cal M}}_{w,\epsilon}(I)$ is: \beq \delta C^{-{
			\cal M}}_{w,\epsilon}(I) = - \sum_{y\in \mathfrak I} w(x,y) \frac{I(y)}{(I(x)+ I(y))^2}s_\epsilon(I(x)-I(y))
	+ S_\epsilon,
	\eeq
	where $S_\epsilon= O(Q_{2,\epsilon}(I(x)-I(y)))$.
	
	\smallskip \noindent In all cases $Q_{1,\epsilon}(z), Q_{2,\epsilon}(z) \to 0$ as $\epsilon\to 0$, uniformly in $z\in [-1,1]$.

\end{proposition}

\medskip Thus we know that $S_\epsilon = O(\epsilon)$ if $A_\epsilon(z)$ is given by Example a), and
$S_\epsilon = O(\epsilon \log 1/\epsilon)$ if $A_\epsilon(z)$ is given by Example b).
These are the cases of interest for us in the experiments.

\medskip
The proof of this proposition will be given in Appendix VIII-C.

\medskip Notice that, letting $\epsilon\to 0$ we have that $\delta C^{f}_{w,\epsilon}(I) \to \delta C^{f}_{w,0}(I)$, where
\begin{eqnarray*}
	\delta C^{\,\rm id}_{w,0}(I) & = & - \frac{1}{2}\sum_{y\in \mathfrak I} w(x,y)\frac{I(y)}{{\rm max}(I(x),I(y))^2} \, {\rm sign}_0(I(x)-I(y)) \\
	& = &  - \frac{1}{2}\left(\sum_{\{y\in {\mathfrak I}\,:\, I(x) > I(y) \}} w(x,y)  \frac{I(y)}{I(x)^2} \, -
	\sum_{\{y\in {\mathfrak I} \, : \, I(x) < I(y) \}} w(x,y)  \frac{1}{I(y)} \right);
\end{eqnarray*}

\medskip

\begin{eqnarray*}  \delta C^{\,
		\rm log}_{w,0}(I)  & = &  - \frac{1}{2}\sum_{y\in \mathfrak I} w(x,y)\frac{1}{I(x)} \, {\rm sign}_0(I(x)-I(y))
	\\
	& = &  - \frac{1}{2}\left(\sum_{\{y\in {\mathfrak I}\,:\, I(x) > I(y) \}} w(x,y)  \frac{1}{I(x)} \, -
	\sum_{\{y\in {\mathfrak I} \, : \, I(x) < I(y) \}} w(x,y)  \frac{1}{I(x)}\right);
\end{eqnarray*}

\medskip

\begin{eqnarray*} \delta C^{\,
		-{\cal M}}_{w,0}(I) & = &  - \sum_{y\in \mathfrak I} w(x,y)\frac{I(y)}{(I(x) + I(y))^2} \, {\rm sign}_0(I(x)-I(y))
	\\
	& = &  - \sum_{\{y\in {\mathfrak I}\,:\, I(x) > I(y) \}} w(x,y)  \frac{I(y)}{(I(x) + I(y))^2} +
	\sum_{\{y\in {\mathfrak I} \, : \, I(x) < I(y) \}} w(x,y)  \frac{I(y)}{(I(x) + I(y))^2}.
\end{eqnarray*}

Now, by direct computation, we have that the derivative of the entropic dispersion term is:
\beq\label{prop:disp_entrop}
\delta D_{\alpha,\beta}^{\cal E}(I) =
\alpha\left(1-\frac{1}{2I(x)}\right) + \beta
\left(1-\frac{I_0(x)}{I(x)}\right).\eeq
We can see that this
expression has a degree of homogeneity -1 with respect to $I(x)$,
the same as the variation of the three contrast terms $C^f_w(I)$.

Assume that $\alpha,\beta > 0$ are fixed. If $E^{f}_{w,\epsilon}(I) = D_{\alpha,\beta}^{\cal E}(I) +  C_{w,\epsilon}^{f}(I)$, $f={\rm id}, \, \log, \, -{\cal M}$, then by linearity of the variational derivative, we have
$\delta E^{f}_{w,\epsilon}(I)=\delta D_{\alpha,\beta}^{\cal E}(I) +
\delta C_{w,\epsilon}^{f}(I)$.
The minimum of $E^{f}_{w,\epsilon}(I)$ satisfies $\delta E^{f}_{w,\epsilon}(I) = 0.$
To search for the minimum we use a semi-implicit discrete gradient descent strategy with respect to
$\log I$. The continuous gradient descent equation is
\beq\label{eq:gradescminmax}
\partial_t \log I = -\delta E^{f}_{w,\epsilon}(I), \eeq
being $t$ the evolution parameter. Since $\partial_t \log
I = \frac{1}{I}\partial_t I$, we have
\beq \label{eq:cont_grad_descA}
\partial_t I = - I \delta E^{f}_{w,\epsilon}(I).
\eeq
Using the gradient descent in $\log I$ leads to (\ref{eq:cont_grad_descA}), which
is related to a gradient descent approach which uses \emph{the relative entropy as a metric}, instead of the
usual quadratic distance (see \cite{Ambrosio:05}).

Let us now discretize our scheme: choosing a finite evolution step $\Delta t
>0$ and setting $I^{k}(x) = I_{k\Delta t}(x)$, $k=0,1,2,\ldots$, being $I^0(x)$ the
original image, thanks to (\ref{prop:disp_entrop}), we can write the semi-implicit discretization of
(\ref{eq:cont_grad_descA}) as
\begin{equation}
	\frac{I^{k+1}(x)-I^k(x)}{\Delta t}  =
	\alpha\left(\frac{1}{2}-I^{k+1}(x)\right) + \beta
	\left(I_0(x)-I^{k+1}(x)\right) - I^k(x) \delta C_{w,\epsilon}^{\, f}(I^k).
\end{equation}
Now, considering the explicit expressions of $\delta C_{w,\epsilon}^{\, f}(I^k)$, neglecting their second terms containing $S_\epsilon$
(see Proposition \ref{prop:var_minmax} ($i$),($ii$),($iii$)) and
performing some easy algebraic manipulations, we find the equation

\beq\label{eq:race}
I^{k+1}(x) = \frac{I^k(x) + \Delta t \left( \frac{\alpha}{2} +
	\beta I_0(x) + \frac{1}{2} R^{\, f}_{\epsilon,I^k}(x)\right)}{1+\Delta t(\alpha + \beta)},
\eeq
where the function $R^{\, f}_{\epsilon,I^k}(x)$ assumes three different forms for $f={\rm id}, \, \log, \, -{\cal M}$, precisely
\beq \label{eq:rid}
R^{\, \rm id}_{\epsilon,I^k}(x) := - 2I^k \delta C_{w,\epsilon}^{\rm id}(I^k) = \sum_{y\in \mathfrak I} w(x,y)
\frac{I^k(x)I^k(y)}{{\rm max}_\epsilon(I^k(x),I^k(y))^2}s_\epsilon(I^k(x)-I^k(y)).
\eeq

\medskip

\beq \label{eq:rlog} R^{\, \log}_{\epsilon, I^k}(x) := - 2I^k \delta C_{w,\epsilon}^{\rm log}(I^k) =
\sum_{y\in \mathfrak I} w(x,y) \, s_\epsilon(I^k(x)-I^k(y)).
\eeq

\medskip

\beq \label{eq:rm} R^{\, \cal M}_{\epsilon,I^k}(x) := - 2I^k \delta C_{w,\epsilon}^{-{\cal M}}(I^k) = \sum_{y \in \mathfrak I} w(x,y) \,
\frac{2I^k(x) I^k(y)}{ ( I^k(x) + I^k(y) )^2} s_\epsilon(I^k(x)-I^k(y)).
\eeq

At the limit $\epsilon \to 0$ we have:

\beq \label{eq:rid0}
R^{\, \rm id}_{0,I^k}(x) = \sum_{\{y\in {\mathfrak I}\,:\, I^k(x) > I^k(y) \}} w(x,y)  \frac{I^k(y)}{I^k(x)} -
\sum_{\{y\in {\mathfrak I} \, : \, I^k(x) < I^k(y) \}} w(x,y)  \frac{I^k(x)}{I^k(y)};
\eeq

\medskip

\beq
\label{eq:rlog0} R^{\, \log}_{0, I^k}(x) = \sum_{y\in \mathfrak I} w(x,y) \, \sgn_0(I^k(x)-I^k(y));
\eeq

\medskip

\beq \label{eq:rm0} R^{\, -\cal M}_{0,I^k}(x) = \sum_{\mathfrak I} w(x,y) \,
\frac{2I^k(x) I^k(y)}{ ( I^k(x) + I^k(y) )^2} \, \sgn_0(I^k(x)-I^k(y)).
\eeq

\medskip

In Appendix VIII-D,
we will comment on the \emph{stability} of these iterative methods.

\section{Features of the three computational algorithms and comparison with existing models}\label{sec:analysis}

\begin{figure}[!h]
	\centering
	\includegraphics[width=2in]{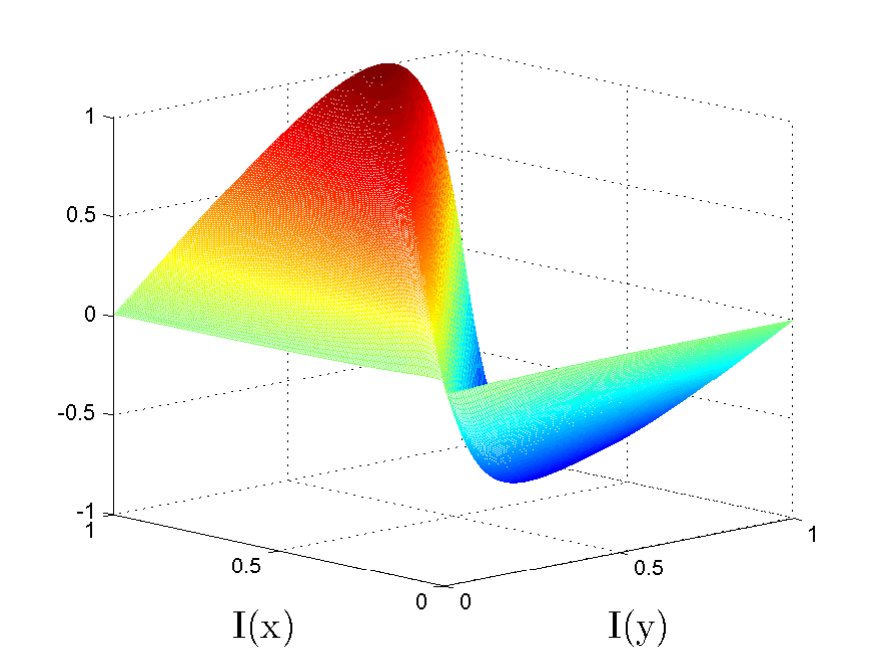}\,
	\includegraphics[width=2in]{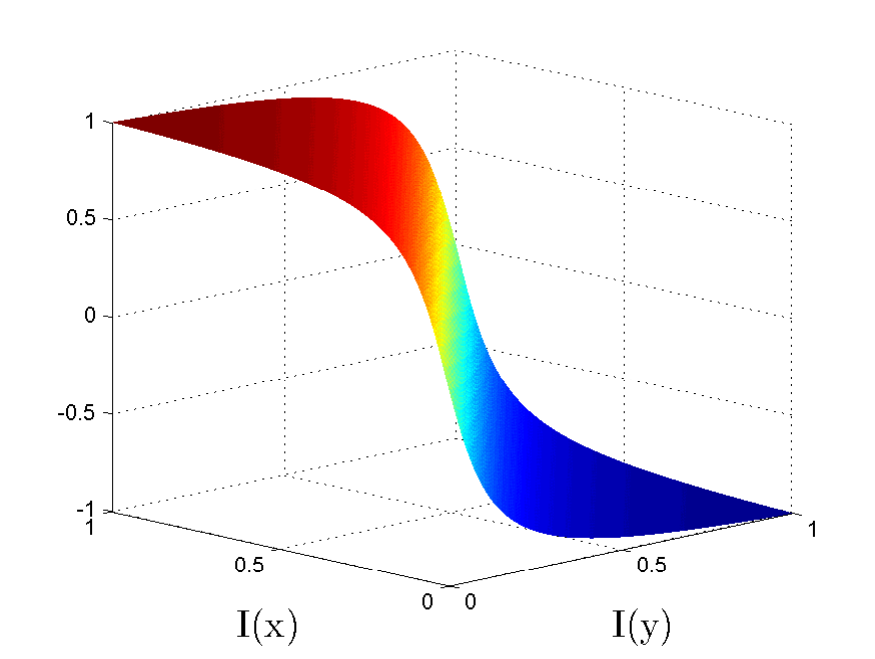} \,
	\includegraphics[width=2in]{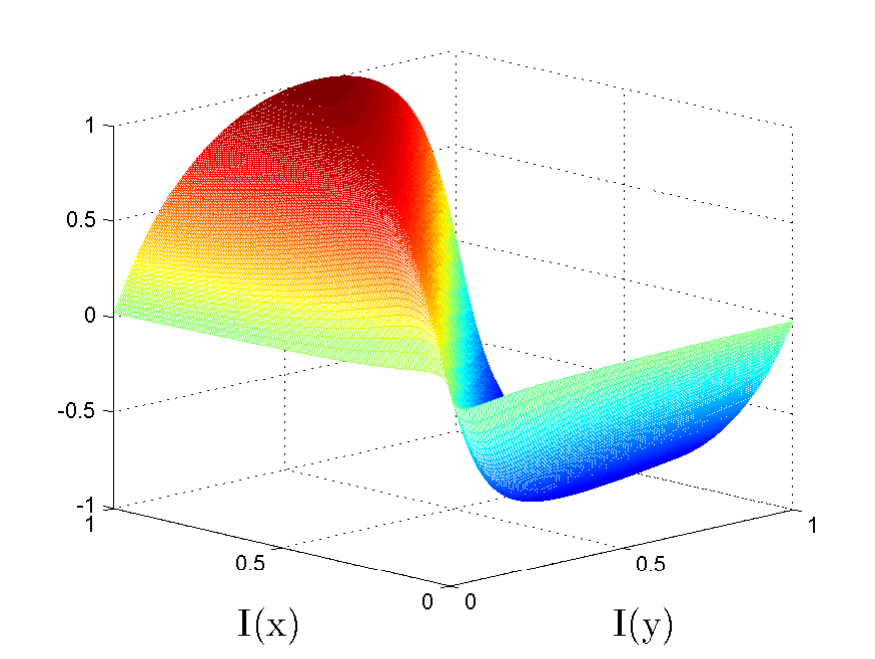}
	\caption{{\em Left}: Surface of $r^{\rm id}_\epsilon$. {\em Center}: Surface of $r^{\log}_\epsilon$. {\em Right}: Surface of $r^{-\cal M}_\epsilon$. Corresponding to the function $s_\epsilon(z) = \arctan(z/\epsilon)/\arctan(1/\epsilon)$ with $\epsilon = 1/20$.}\label{fig:superficiess}
\end{figure}

Equation (\ref{eq:race}) can be used to implement three iterative
computational algorithms for color image enhancement, as $f$ varies. Notice that the $R$ functions share
an identical structure, i.e. $R^f_{\epsilon,I^k}(x)=\sum_{y\in \mathfrak I} w(x,y) \, r^f_\epsilon(I^k(x),I^k(y))$, for a suitable function $r^f_\epsilon$. In Fig. \ref{fig:superficiess} we show the corresponding
surfaces.

We can see that the surfaces representing $r_\epsilon^{\rm id}$ and
$r_\epsilon^{- \cal M}$ are very similar, the only difference being that the
second is quite smoother than the first. Hence, we expect the
corresponding algorithms to have similar enhancing properties.

Instead, the surface representing $r_\epsilon^{\log}$ shows an opposite
behavior with respect to the previous two surfaces at the extreme
points. Moreover, since the logarithmic function grows very
rapidly for small values of its argument and slowly for higher
ones, we expect the corresponding algorithm to perform a strong
contrast enhancement in dark image areas and a weak one in bright
zones. In section \ref{sec:tests} it will be shown that all these
observations are verified by empirical tests.

We now want to show that there is a precise correspondence between
the algorithm corresponding to $E^{\log}_{w,\epsilon}$ and ACE
\cite{Rizzi:03}, a recent perceptually inspired color correction
algorithm, while the algorithms corresponding to the energies
$E^{\rm id}_{w,\epsilon}$, $E^{-\cal M}_{w,\epsilon}$ are
\emph{qualitatively} related with the Retinex model, but they also
exhibit new peculiarities that distinguish them from Retinex.

\subsection{Correspondence between $E^{\log}_{w,\epsilon}$ and the ACE model}

The original ACE model has been developed as a perceptually
inspired color correction algorithm based on the already mentioned
`gray-world principle', i.e. that the average color of a scene is
the middle gray \cite{Rizzi:03}. The similarities between the ACE equations
and those relative to variational histogram equalization
\cite{Sapiro:97} lead to a variational
formulation of ACE and a deeper understanding of the model
\cite{Bertalmio:07}. The functional energy of ACE is \cite{Bertalmio:07}: \beq
E^{\rm ACE}_{w}(I):=  \sum_{x\in \mathfrak I} \left[\frac{\alpha}{2}
\left(I(x)-\frac{1}{2}\right)^2 + \frac{\beta}{2}
\left(I(x)-I_0(x)\right)^2 \right] - \frac{1}{4}
\sum_{x\in \mathfrak I} \sum_{y\in \mathfrak I} w(x,y) \, {\rm S}(I(x)-I(y)), \eeq
being S the primitive function of a sigmoid s$:[-1,1] \to [-1,1]$ and $\alpha,\beta > 0$.

Now, applying a semi-implicit gradient descent technique to minimize $E^{\rm
	ACE}_{w}$ with respect to $I$, one arrives to the
iterative scheme:
\beq I^{k+1}(x) =
\frac{I^k(x) + \Delta t \left( \frac{\alpha}{2} + \beta I_0(x) + \frac{1}{2} R^{\, \rm
		ACE}_{I^k}(x)\right)}{1+\Delta t (\alpha+\beta)},
\eeq
where
\beq\label{eq:ace_basico} R^{\, \rm
	ACE}_{I^k}(x) := \sum_{y\in \mathfrak I} w(x,y) \, {\rm s}(I^k(x) -
I^k(y)).
\eeq
Since the sigmoid can be considered as an approximation of the sign function,
the algorithms defined by (\ref{eq:rlog}) and (\ref{eq:ace_basico}) are essentially identical.
We recall that the sigmoid in ACE has been used instead of the signum simply to avoid abrupt color corrections.

\subsection{Qualitative similarities and differences between $E^{\rm id}_{w,\epsilon}$, $E^{-{\cal M}}_{w,\epsilon}$ and the Retinex model}

The models represented by the energies $E^{\, \rm id}_{w,\epsilon}$, $E^{-{ \cal
		M}}_{w,\epsilon}$ are, to the authors' knowledge, novel in the literature.
As we will show in the experiments in Section \ref{sec:tests}, the output images of their corresponding
computational algorithms are very similar. Analytically (modulo replacing the function ${\rm sign}_\epsilon$ by ${\rm sign}_0$
for the sake of interpretation), this is
motivated by the fact that both $R_{\epsilon,I^k}^{\, \rm id}(x)$ and
$R_{\epsilon,I^k}^{\, -\cal M}(x)$ are represented by integrals of intensity
ratios bounded by 1 taken with positive sign when $I(x)> I(y)$ and
negative sign in the opposite situation. The presence of ratios
bounded by 1 is one of the features of the original Retinex model
\cite{Land:71}. In fact, as it has been proven in
\cite{Provenzi:05}, the peculiar `ratio-reset' Retinex
computation can be equivalently represented as the ratio between
the intensity of any fixed pixel and the highest intensity sampled
by a geometric structure used to extract spatial chromatic
information around that pixel. The geometric structures used in the
various Retinex implementations are many: paths, masks, sprays,
etc. See, e.g., \cite{Provenzi:07} for a more detailed overview.

The fundamental difference between Retinex and equations
(\ref{eq:rid}) and (\ref{eq:rm}) is that, in the
Retinex model, ratios are performed only over the highest intensity
value and they are always taken with positive sign. As a
consequence, as proved in \cite{Provenzi:05}, the ratio-reset Retinex operation is
unbalanced, i.e. it always increases the intensity of any pixel (a post-LUT processing can overcome this problem, but this operation is not intrinsically contained in the ratio-reset Retinex equations).
Instead, equations (\ref{eq:rid}) and (\ref{eq:rm})
can both increase or reduce pixel intensity. For this reason, the
energies $E^{\, \rm id}_{w,\epsilon}$, $E^{-{ \cal
		M}}_{w,\epsilon}$, could be \emph{qualitatively} thought as
describing a `symmetrized Retinex model'. Later we will compare
the action of this model with the one of Retinex on an overexposed
image, exhibiting the advantages provided by the symmetrization just commented.
Finally, we would like to mention that a heuristic attempt to unify the qualities of Retinex and those of ACE in a single algorithm called RACE has been made in \cite{Provenzi:08}, equations (\ref{eq:rid}) and (\ref{eq:rm}) can be thought as a rigorous variational generalization of the RACE algorithm.

\section{A general strategy for computational complexity reduction}

The computational complexity of the algorithms derived from the minimization of the
energy functionals that we have examined is ${\cal O}(N^2)$,
$N$  being the number of pixels of the input image. In fact, the computational
complexity of $R^{\, id}_{\epsilon,I^k}(x)$, $R^{\, \log}_{\epsilon,I^k}(x)$ and
$R^{\, -\cal M}_{\epsilon,I^k}(x)$ is of order ${\cal O}(N)$ for every $x$,
hence the \emph{global} computational complexity is of order ${\cal
	O}(N^2)$. This implies that with a standard PC (P4, 3GHz) it can
take up hours to process a high resolution image.

In \cite{Provenzi:07} a local sampling technique has been devised to reduce time computation. Here we provide a different technique, based on mathematical properties of the $r$-functions rather that on sampling. This permits to maintain all the chromatic information, avoiding sampling noise.

From now on, we assume that $w$ is a function of $\Vert x-y\Vert_{\mathfrak{I}}$ and
we show how to speed up the computation of $R^{f}_{\epsilon,I^k}(x)$, reducing its
computational complexity to ${\cal O}(N\log N)$ by means of a suitable
approximation, which generalizes an analogous technique used in
\cite{Bertalmio:07} to speed up the ACE algorithm. Let us denote with $R(x)=\sum_{y\in \mathfrak I} w(x,y) \, r(I(x),I(y))$
any of the three $R$-functions specified in equations (\ref{eq:rid}), (\ref{eq:rlog}), (\ref{eq:rm}), omitting the indexes $f$, $\epsilon$ and $I^k$ both to simplify the notation and because their specification is not important for what follows.

The basic observation underlying our proposal is that, if we manage to separate the dependence of
the function $r$ on $I(x)$ and $I(y)$, then we can express $R(x)$
as a sum of convolutions. In fact, let us suppose that $R(x)$
admits a polynomial approximation $R^{(n)}(x)$, being $n\in \mathbb N$ the
approximation degree, such as
\beq\label{eq:approx}
R^{(n)}(x)=\sum_{j=0}^n f_j(I(x)) \sum_{y \in \mathfrak I} w(x,y) \,
g_j(I(y)) \, ,
\eeq
where $f_j$ are functions of $I(x)$ and
$g_j$ are functions of $I(y)$. Since $w$ is a function of
$x-y$, it is clear that the integrals appearing in
(\ref{eq:approx}) actually reduce to convolutions, which can be
computed using the Fast Fourier Transform
(FFT), whose computational complexity is of
order ${\cal O}(N\log N)$. Hence, if $n<<N$, the computational
complexity of the approximated algorithm reduces to ${\cal
	O}(N\log N)$.

To achieve the separation of variables in (\ref{eq:approx}), we fix an
approximation degree $n$ and we use a numerical algorithm to find the
polynomial $p(I(x),I(y))=\sum_{j=0}^n \sum_{l=0}^j p_{j-l,l} \,
I(x)^{j-l} \, I(y)^{l}$, $p_{j-l,l}\in \re$, that minimizes the
$L^2$ distance to $r(I(x),I(y))$. Indeed, since $r$ is computed only once (before the first gradient descent iteration,) we can choose the degree $n$
to control the maximum error. The polynomial can be easily
rearranged and expressed as $p(I(x),I(y))=\sum_{j=0}^n f_j(I(x))
\, I(y)^{j}$ for suitable functions $f_j(I(x))$. This expression can be directly used in equation
(\ref{eq:approx}) setting $g_j(I(y))\equiv I^j(y)$ for every
$j=1,\ldots,n$. Increasing $n$, one can reduce the approximation
error; our experiments show that for the functions $r_\epsilon^{\rm id}$,
$r_\epsilon^{\log}$ and $r_\epsilon^{-\cal M}$, the value $n=9$ gives a
satisfactory approximation, so we have used it in all of our tests.

For the sake of completeness, we notice that, if the surface
representing the function $r(I(x),I(y))$ has discontinuities or strong gradients,
a polynomial expansion may not be useful for its
approximation. In such a case, a Fourier expansion could be more
suitable than the polynomial one, but the corresponding expression
of $g_j(I(y))$ would be analytically more complicated, being, in
this case, expressed in terms of trigonometric functions instead
of powers.

\section{Tests}\label{sec:tests}

So far, we have performed a theoretical investigation about the
proposed axiomatic framework for perceptually inspired variational
color enhancement. In this section, we show and discuss some tests
about the algorithms derived from the three functional energies
examined. We will work on images acquired with generic digital cameras.
Before going into details, it is worthwhile to recall
that the algorithms represented by equations (\ref{eq:race}), (\ref{eq:rid}),
(\ref{eq:rlog}) and (\ref{eq:rm}) have four
explicit parameters: $\alpha$, $\beta$, $\epsilon$ and $\Delta t$, moreover,
the function $w$ determines the local contrast enhancement behavior. We have chosen the function $w(x,y) = \frac{A}{\Vert x - y\Vert_{\mathfrak{I}}}$, where $A$ is the normalization parameter, so that (\ref{eq:norma}) is satisfied.
Recall that $\alpha$ controls the dispersion around the middle gray, $\beta$
sets the strength of attachment to original data, and $\Delta t$ is the
gradient descent step. The parameter $\epsilon$ controls the regularization of the signum
function. In practice, we have chosen $s_\epsilon(z) = \arctan(z/\epsilon)/\arctan(1/\epsilon)$
because the signum function is too singular to be used without leading to unwanted noise amplification.

Our tests indicate that an overall satisfactory set of parameters
is $\alpha=\frac{255}{253}$, $\beta=1$, $\Delta t=0.2$, and $\epsilon = \frac{1}{20}$. We iterate until the average Mean Square Error (MSE) per pixel (between the images in the current and previous iteration) falls below a threshold of $10^{-4}$ (typically the steady state is reached within 10-20 iterations.)
Varying these parameters with respect to image characteristics could possibly give better results. We consider the
image-dependency of parameters an interesting issue to be investigated, but its thorough and significative analysis falls outside the scope of this paper.

\subsection{Some experimental results}

To show the abilities of the computational algorithms that we are
discussing, we consider the three different images shown in
fig. \ref{fig:originals} (courtesy of P. Greenspun).

\begin{figure}[!ht] \label{fig:originals}
	\centering
	\includegraphics[width=1.6in]{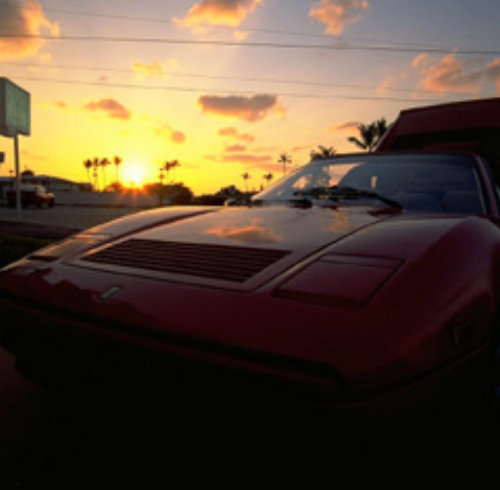}\,
	\includegraphics[width=1.6in]{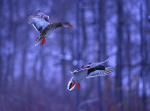}\,
	\includegraphics[width=1.6in]{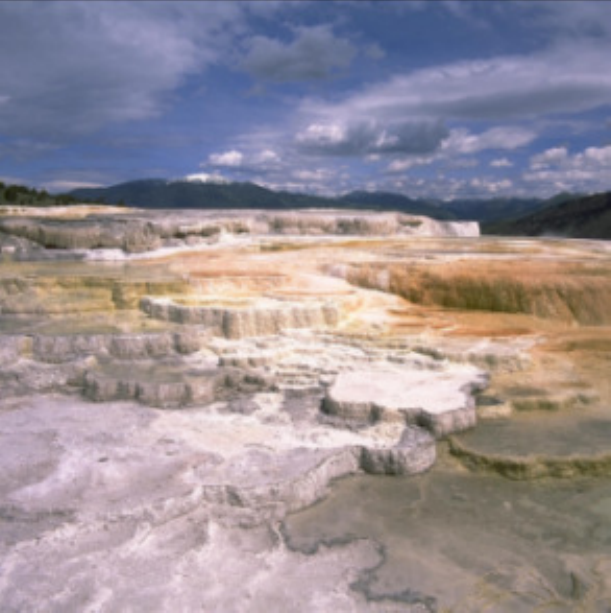}
	\caption{Three different original images of our test set showing different features to be enhanced.}\label{fig:originals}
\end{figure}

\begin{figure}[!ht]
	\centering
	\includegraphics[width=1.6in]{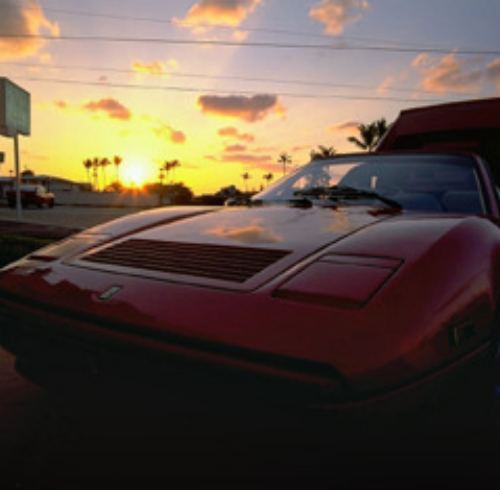}\,
	\includegraphics[width=1.6in]{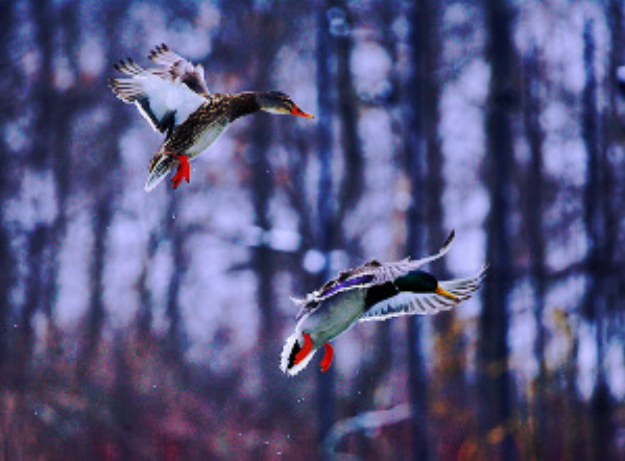}\,
	\includegraphics[width=1.6in]{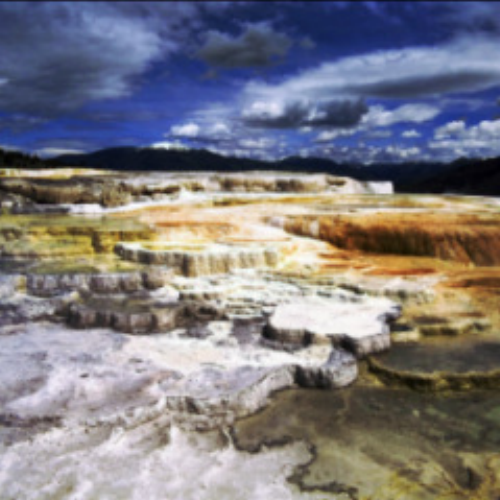}
	\caption{Outputs of the algorithm corresponding to the energy $E^{\, \rm id}_{w,\epsilon}$.}\label{fig:salida_race}
\end{figure}

\begin{figure}[!ht]
	\centering
	\includegraphics[width=1.6in]{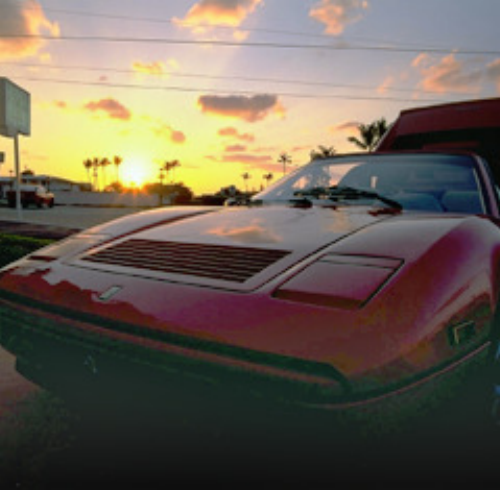}\,
	\includegraphics[width=1.6in]{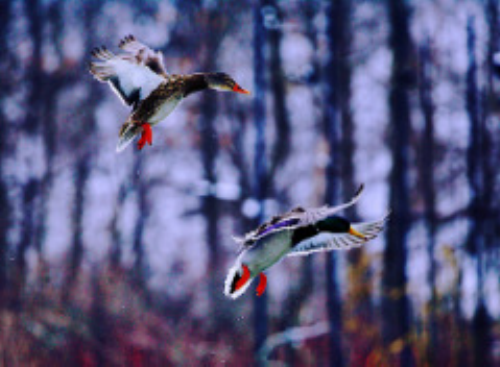}\,
	\includegraphics[width=1.6in]{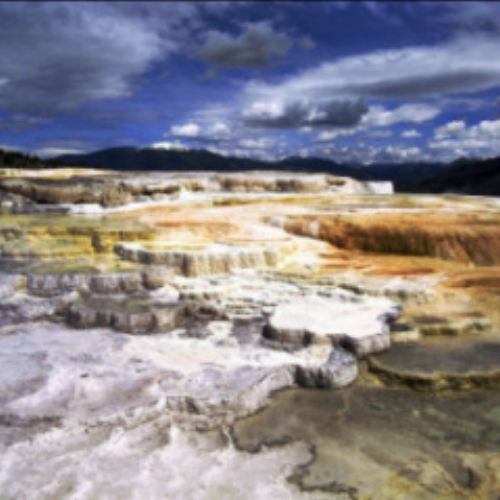}
	\caption{Outputs of the algorithm corresponding to the energy $E^{\, \log}_{w,\epsilon}$.}\label{fig:salida_race_log}
\end{figure}

\begin{figure}[!ht]
	\centering
	\includegraphics[width=1.6in]{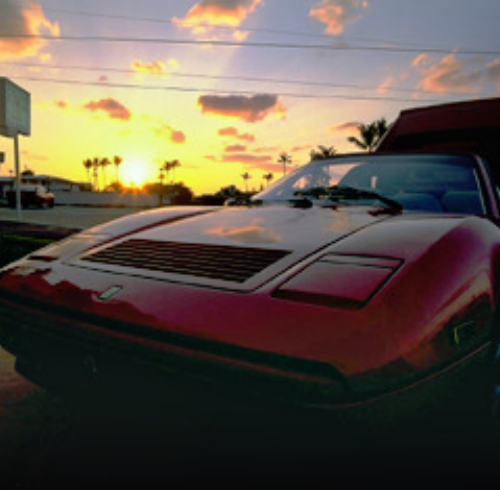}\,
	\includegraphics[width=1.6in]{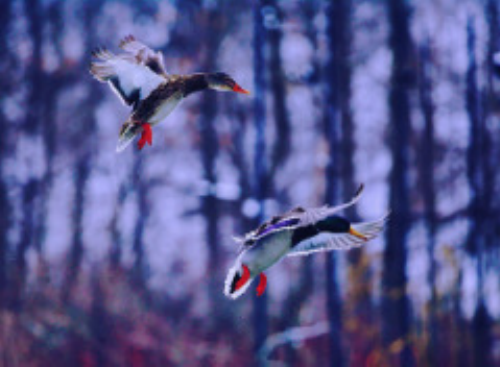}\,
	\includegraphics[width=1.6in]{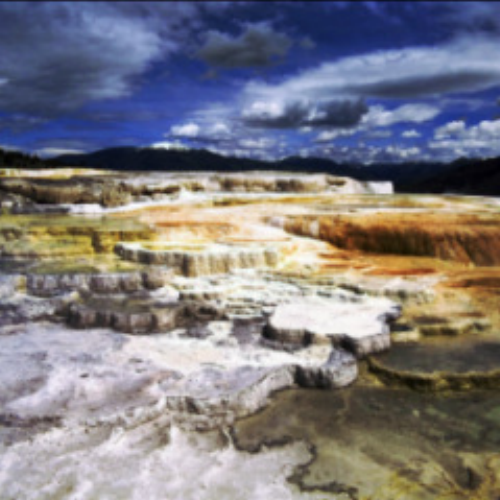}
	\caption{Outputs of the algorithm corresponding to the energy $E^{\, -\cal M}_{w,\epsilon}$.}\label{fig:salida_race_michelson}
\end{figure}

The image on the left is underexposed, the image in the center has
a strong blue cast, and the one on the right is quite overexposed.
The results of the algorithms corresponding to equations
(\ref{eq:rid}), (\ref{eq:rlog}) and (\ref{eq:rm})
are presented in figs. \ref{fig:salida_race},
\ref{fig:salida_race_log} and \ref{fig:salida_race_michelson},
respectively.

Tests confirm that the considered algorithms indeed enhance
contrast, remove color cast and equalize both underexposed and
overexposed pictures. As expected, the algorithms derived from
energies $E^{\, \rm id}_{w,\epsilon}$ and $E^{- {\cal
		M}}_{w,\epsilon}$ have very similar enhancement
characteristics, while they differ from the one based on the energy
$E^{\, \log}_{w,\epsilon}$ in the treatment of low and high gray
levels. This latter algorithm has greater contrast enhancement
properties in dark regions, but the ones corresponding to
$E^{\, \rm id}_{w,\epsilon}$ and $E^{-{\cal
		M}}_{w,\epsilon}$ perform better in bright areas.

\begin{figure}[!ht]
	\centering
	\includegraphics[width=1.6in]{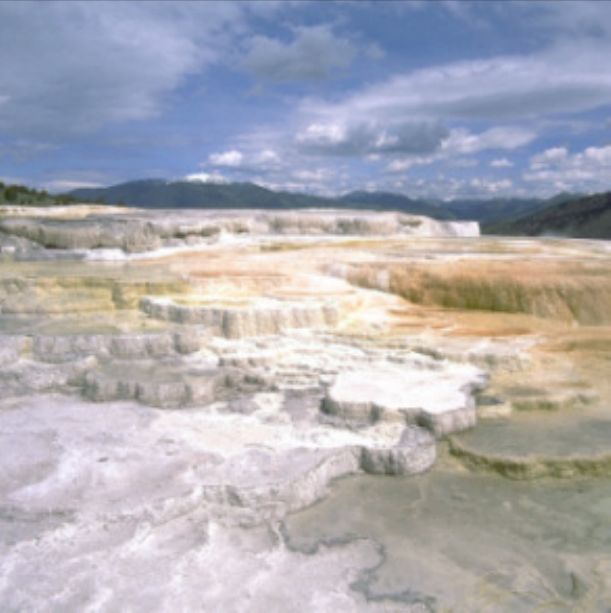}\,
	\includegraphics[width=1.6in]{./images/islanda_race}\,
	\includegraphics[width=1.6in]{./images/islanda_race_michelson}\,
	\caption{{\em Left.} Action of Retinex (in the RSR implementation \cite{Provenzi:07}).
		{\em Center.} Action of the algorithm corresponding to (\ref{eq:rid}).
		{\em Right.} Action of the algorithm corresponding to (\ref{eq:rm}).} \label{fig:islanda_rsr}
\end{figure}

Moreover, regarding the treatment of overexposed images, it is
interesting to compare the effect of the `pure' ratio-reset Retinex
action (without post-LUT procedures) with the results of the algorithms relative to formulas
(\ref{eq:rid}) and (\ref{eq:rm}) on the image shown
in fig. \ref{fig:originals} (right). In fig. \ref{fig:islanda_rsr}
it can be noticed that the result of Retinex is a further
overexposed image, while the other two algorithms are able to
equalize the image.

\subsection{Gamma transformation of the basic contrast variable}

As just commented, the algorithms based on the energies $E^{\, \rm
	id}_{w,\epsilon}$ and $E^{-{\cal M}}_{w,\epsilon}$ have a good behavior
in bright areas, but it would be desirable to increase their
contrast enhancement ability in dark ones. In order to do that,
we have modified the corresponding contrast terms by adding a
gamma transformation. Non-linear operations as gamma
transformations are also performed by the HVS \cite{Pratt:07}. Mannos and Sakrison \cite{Mannos:74} have established that a power law transformation of the form ${I_{\rm in}(x,y)}^\gamma$,
being $I_{\rm in}(x,y)$ the retinal signal and $\gamma \simeq
1/2$, provides good agreement with experimental evidence. Precisely,
chosen a $\gamma \in (0,1)$, we have re-defined the basic contrast
variable as
$\left(\frac{\min(I(x),I(y))}{\max(I(x),I(y))}\right)^\gamma$.
Replacing the basic contrast regularized variable with the gamma-transformed
one, we get new energy functionals that still comply with all
our assumptions, being the gamma transformation strictly monotone.
Explicitly, we redefine the energies $E^{\, \rm
	id}_{w,\epsilon}$ and $E^{-{\cal M}}_{w,\epsilon}$ by changing the contrast terms
$C^{\, \rm id}_{w,\epsilon}$ and $C^{-{\cal M}}_{w,\epsilon}$ by the
gamma-transformed ones. Using the properties of the logarithmic
function, it is easy to see that the consequence of the gamma
transformation on $C_w^{\, \log}(I)$ simply amounts to an overall
coefficient change in the corresponding $R$-function:
$R_{\epsilon,\gamma,I^k}^{\, \log}(x) := \gamma\sum_{y\in {\mathfrak I}} w(x,y)
\, s_\epsilon(I(x) - I(y))$, so we will not consider this trivial
case in our discussion.
\beq C^{\, \rm id}_{w,\epsilon,\gamma}(I) =
\frac{1}{4 \gamma}\sum_{x\in {\mathfrak I}} \sum_{y\in {\mathfrak I}} w(x,y) \,
\left(\frac{{\rm min}_\epsilon(I(x),I(y))}{{\rm max}_\epsilon(I(x),I(y))}\right)^\gamma \, ;
\eeq
\beq C^{\, -\cal M}_{w,\epsilon,\gamma}(I) = - \frac{1}{4
	\gamma} \sum_{x\in {\mathfrak I}} \sum_{y\in {\mathfrak I}} w(x,y) \, \frac{A_\epsilon(I(x)^\gamma - I(y)^\gamma)}{I(x)^\gamma + I(y)^\gamma}.
\eeq
By direct computation, it can be seen that the gradient descent
equations corresponding to the gamma-transformed contrast terms
still have the general form represented by equation
(\ref{eq:race}), but now with these new
$R$-functions
\beq R_{\epsilon,\gamma,I^k}^{\, \rm id}(x) :=
\sum_{y\in {\mathfrak I}} w(x,y) \, \left(\frac{{\rm min}_\epsilon(I^k(x),I^k(y))}{{\rm max}_\epsilon(I^k(x),I^k(y))}\right)^\gamma
s_\epsilon(I^k(x)-I^k(y)),
\eeq
\beq R_{\epsilon,\gamma,I^k}^{\, -\cal M}(x) :=
\sum_{y\in {\mathfrak I}} w(x,y) \, \frac{2\, \left({\rm min}_\epsilon(I^k(x),I^k(y))^\gamma \, {\rm max}_\epsilon(I^k(x),I^k(y))^\gamma\right)}{\left({\rm min}_\epsilon(I^k(x),I^k(y))^\gamma + {\rm max}_\epsilon(I^k(x),I^k(y))^\gamma \right)^{2}}  \,
s_\epsilon(I^k(x) - I^k(y)).
\eeq

\begin{figure}[!ht]
	\centering
	\includegraphics[width=2in]{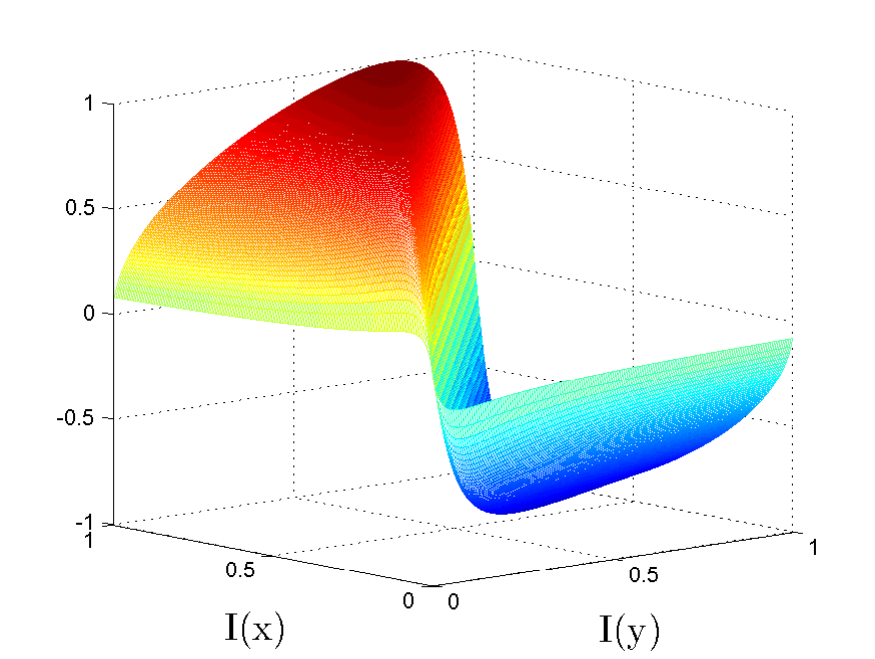}\,
	\includegraphics[width=2in]{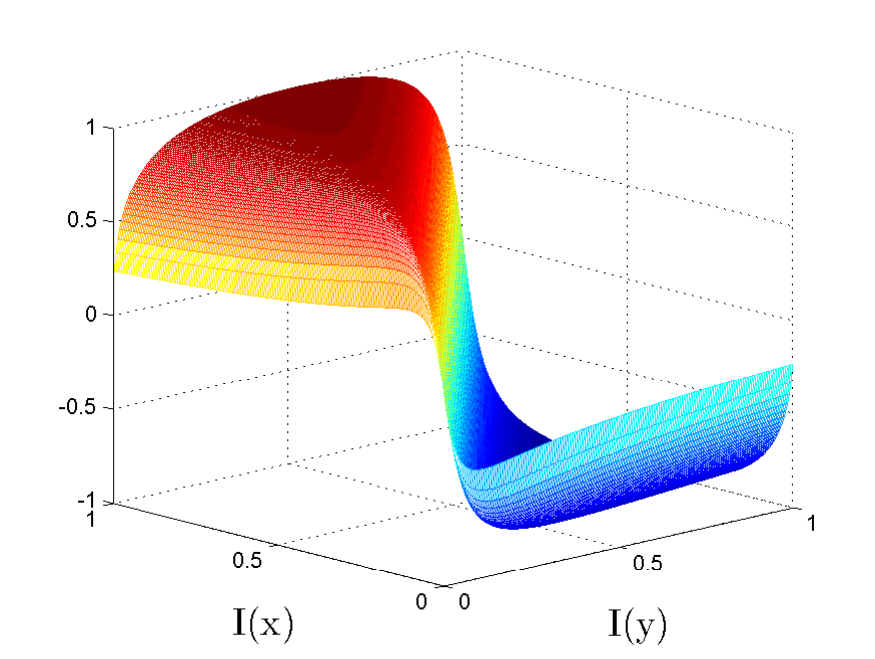}
	\caption{Graphics of $r^{\, \rm id}_{\epsilon,\gamma}$ (left) and $r^{\, -\cal M}_{\epsilon,\gamma}$ (right) with $\gamma = 1/2$ and $s_\epsilon(z) = \arctan(z/\epsilon)/\arctan(1/\epsilon)$ with $\epsilon = 1/20$.}\label{fig:r_gamma}
\end{figure}

The consequences of the gamma transformation on the analytic form
of the $r$-functions $r^{\, \rm id}_{\epsilon,\gamma}(I(x),I(y))$ and
$r^{\, -\cal M}_{\epsilon,\gamma}(I(x),I(y))$, appearing in
$R_{\epsilon,\gamma,I^k}^{\, \rm id}(x)$ and $R_{\epsilon,\gamma,I^k}^{\, - \cal
	M}(x)$, respectively, can be appreciated in fig.
\ref{fig:r_gamma}.

It can be noticed that the gradient of both surfaces around
$(0,0)$ increased, while they both get slightly smoothed for
higher values. The smoothing gets stronger as $\gamma$ decreases,
until the limit $\gamma = 0$, where they reduce to the same
surface, precisely: $r^{\,\log}_\epsilon$. The most suitable value of $\gamma$ depends on the image
content, however our tests showed that an overall good performance with
$\gamma = 1/2$.

\begin{figure}[!ht]
	\centering
	\includegraphics[width=1.6in]{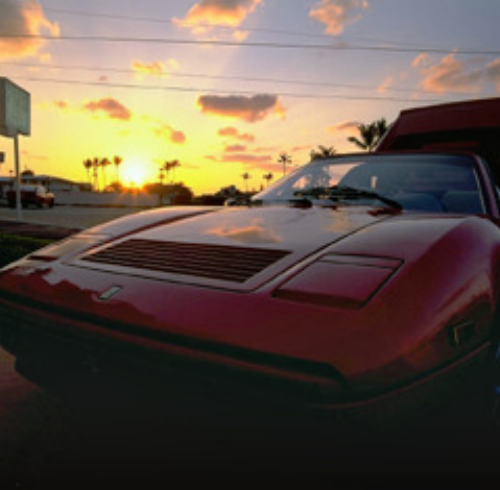}\,
	\includegraphics[width=1.6in]{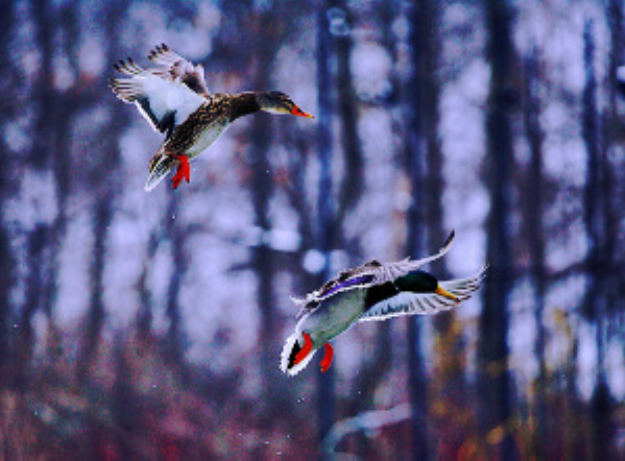}\,
	\includegraphics[width=1.6in]{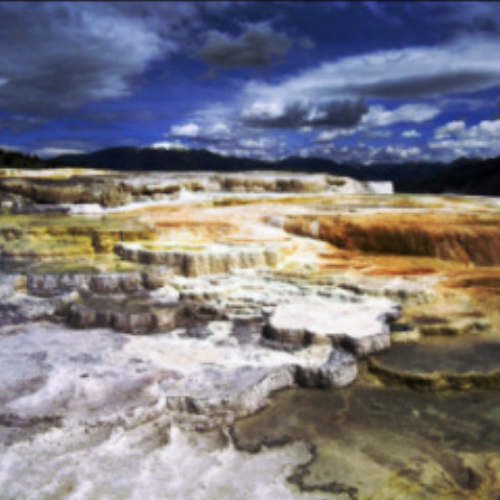}
	\caption{Outputs of the algorithm corresponding to the energy $E^{\, \rm id}_{w,\epsilon}$ with gamma transformed contrast variable, $\gamma=1/2$.}\label{fig:salida_race_gamma}
\end{figure}

\begin{figure}[!ht]
	\centering
	\includegraphics[width=1.6in]{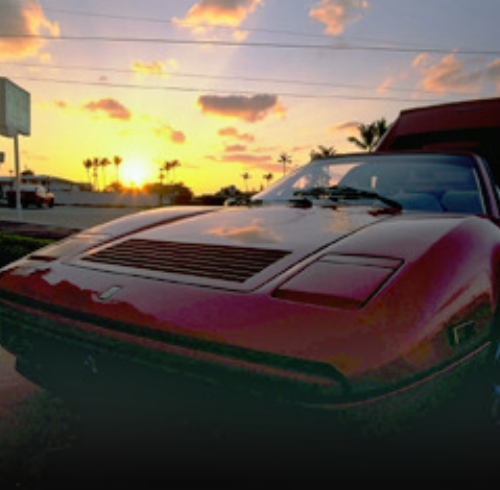}\,
	\includegraphics[width=1.6in]{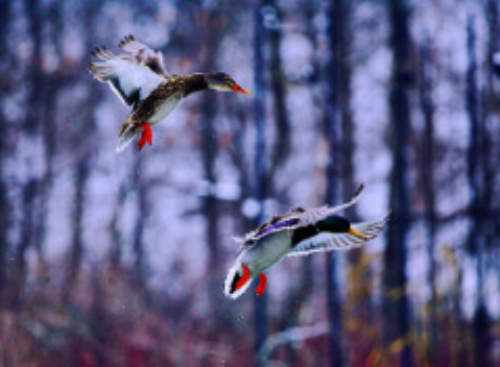}\,
	\includegraphics[width=1.6in]{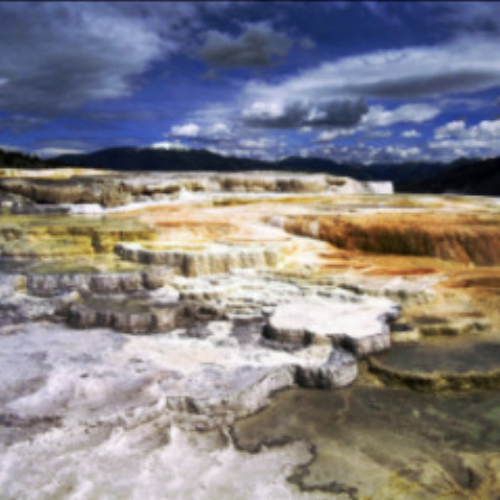}
	\caption{Outputs of the algorithm corresponding to the energy $E^{\, -\cal M}_{w,\epsilon}$ with gamma transformed contrast variable, $\gamma=1/2$.}\label{fig:salida_race_michelson_gamma}
\end{figure}

The pictures in figs. \ref{fig:salida_race_gamma} and
\ref{fig:salida_race_michelson_gamma} confirm that the gamma
transformed algorithms have greater contrast enhancement
properties in dark areas with respect to the original ones, while
maintaining good enhancement properties in bright zones.

We present other examples of results. To save space we present the outputs relative only to the algorithm corresponding to the energy $E^{\, -\rm id}_{w,\epsilon}$ with gamma transformed contrast variable, $\gamma=1/2$.

\begin{figure}[!ht]
	\centering
	\includegraphics[width=1.8in]{./images/book}\,
	\includegraphics[width=1.8in]{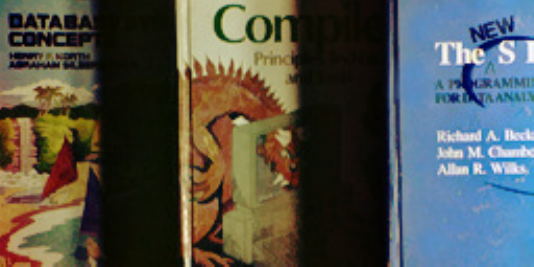} \vskip 0.15cm
	\includegraphics[width=1.8in]{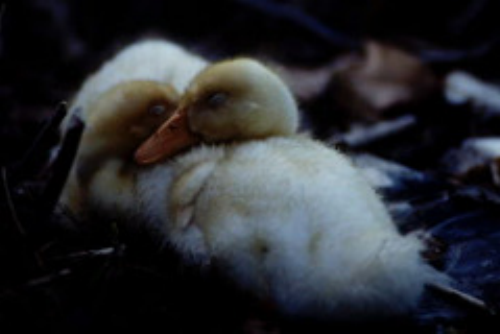}\,
	\includegraphics[width=1.8in]{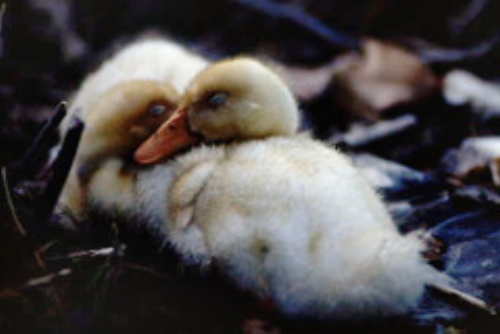} \vskip 0.15cm
	\includegraphics[width=1.3in]{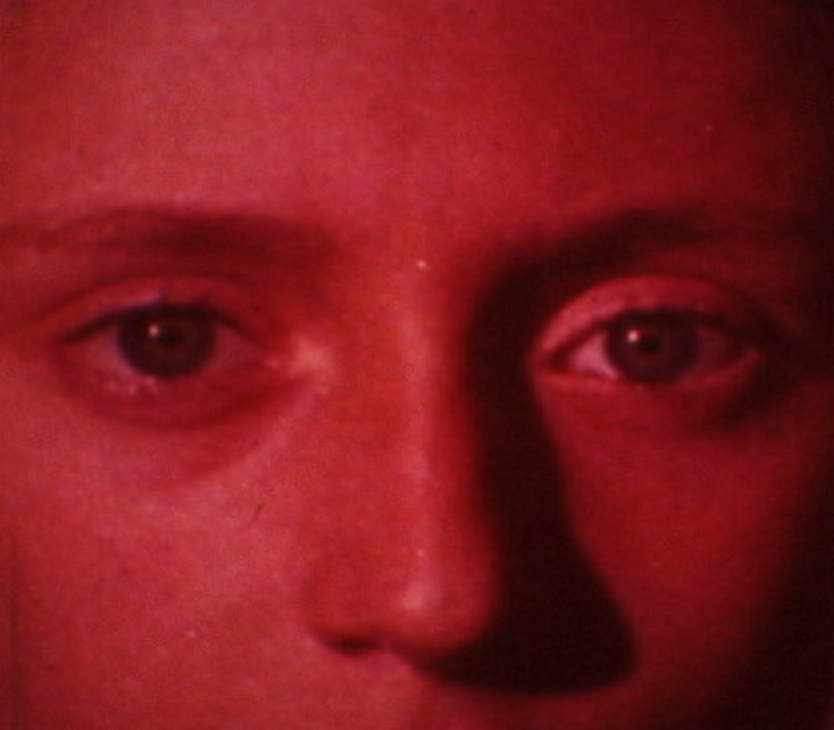}\,
	\includegraphics[width=1.3in]{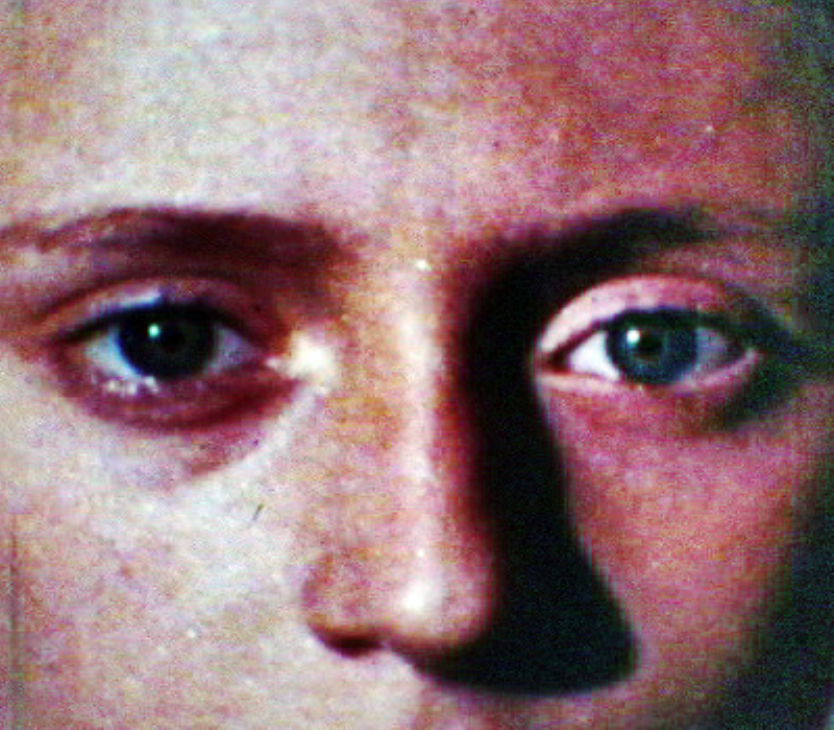} \vskip 0.15cm
	\includegraphics[width=1.3in]{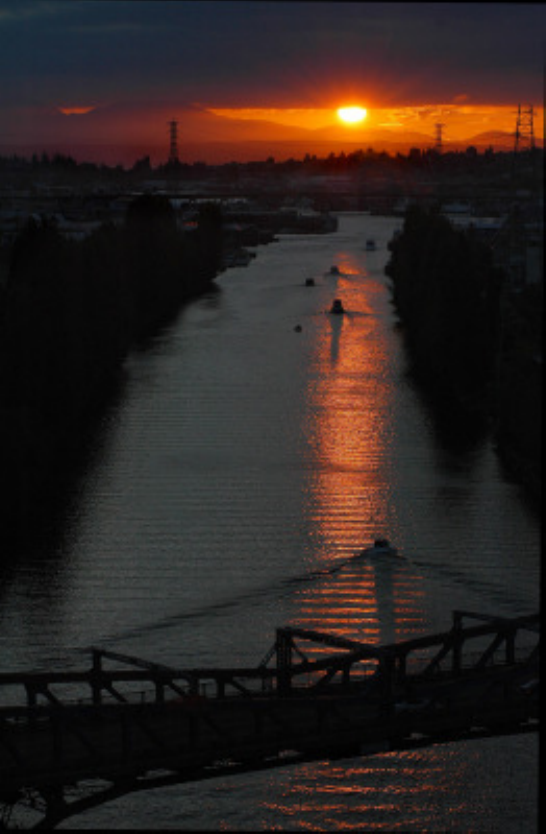}\,
	\includegraphics[width=1.3in]{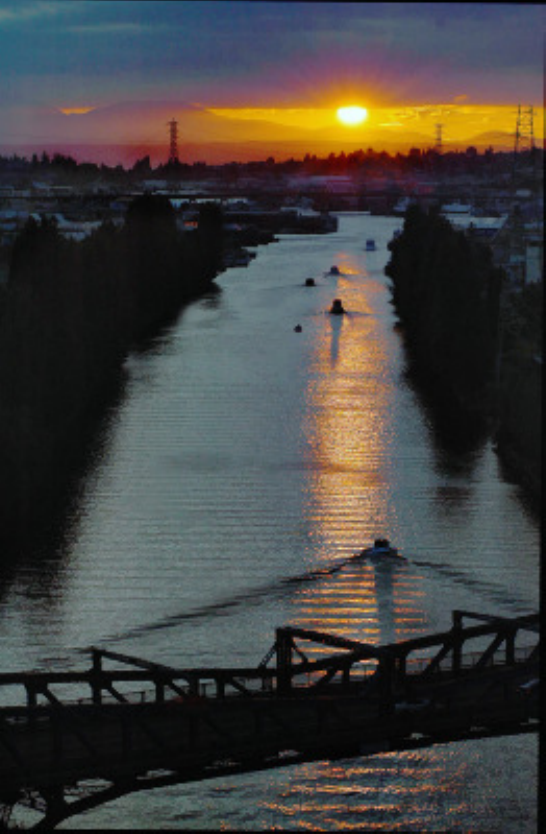}
	\caption{{\em Left}: original images. {\em Right}: filtered versions with the algorithm corresponding to the energy $E^{\, -\rm id}_{w,\epsilon}$ with gamma transformed contrast variable, $\gamma=1/2$.}\label{fig:otras_salidas}
\end{figure}

\subsection{Tests on cognitive images}

\begin{figure}[!ht]
	\centering
	\includegraphics[width=1in]{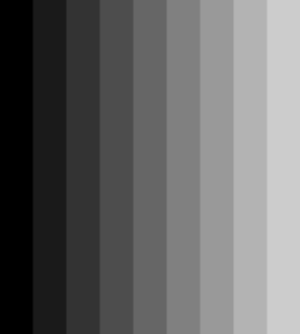}\,
	\includegraphics[width=1in]{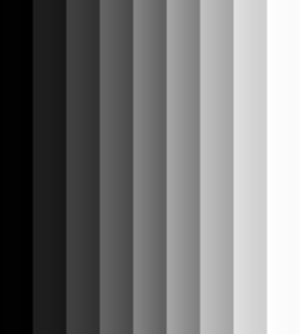}
	\caption{{\em Left}: original Mach band image. {\em Right}: filtered version with the algorithm corresponding to the energy $E^{\, -\rm id}_{w,\epsilon}$ with gamma transformed contrast variable, $\gamma=1/2$.}\label{fig:mach_bands}
\end{figure}

\begin{figure}[!ht]
	\centering
	\includegraphics[width=1.6in]{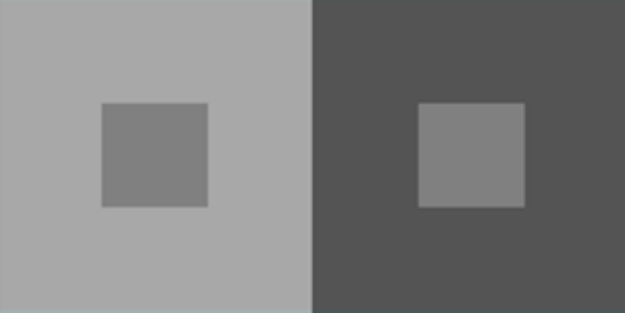}\,
	\includegraphics[width=1.6in]{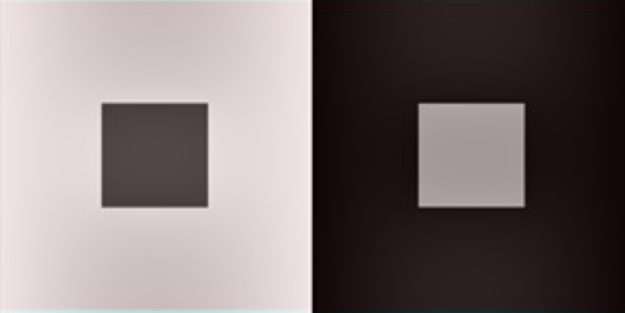}
	\caption{{\em Left}: original simultaneous contrast image. {\em Right}: filtered version with the algorithm corresponding to the energy $E^{\, -\rm id}_{w,\epsilon}$ with gamma transformed contrast variable, $\gamma=1/2$.}\label{fig:contrasto_simultaneidad}
\end{figure}

We have also tested the ability of our algorithms to reproduce optical illusions. Here we present in Figs. \ref{fig:mach_bands} and \ref{fig:contrasto_simultaneidad} our results on two classical images: Mach bands and simultaneous contrast, respectively. As can be seen, the results are coherent with the experimental evidences.

\subsection{A pre and post-processing strategy for noise control}

\begin{figure}[!ht]
	\centering
	\includegraphics[width=1.6in]{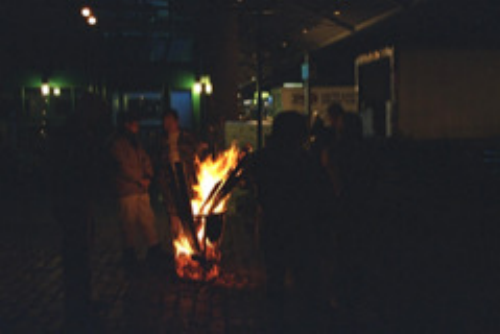}\,
	\includegraphics[width=1.6in]{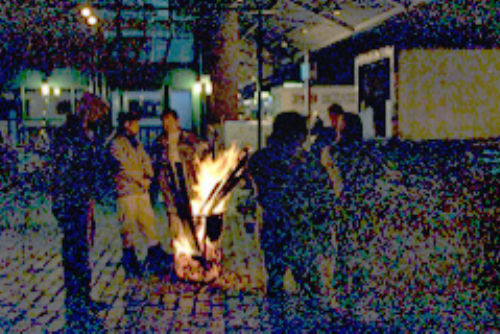}\,
	\includegraphics[width=1.6in]{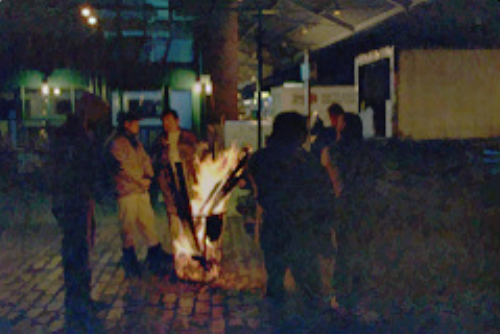}
	\caption{\emph{Left:} A low key image. \emph{Center:} its filtered version after the application of the algorithm derived by $E^{\, \rm id}_{w,\epsilon}$ (the
		algorithms relative to the other energies produce the same effect). \emph{Right:} noise control provided by the strategy exposed in the text.}\label{fig:fuoco}
\end{figure}

Even though the algorithms we are discussing do not
\emph{introduce} noise in images, they can \emph{enhance} noise if it is already present. In particular, there can be `limit'
situations in which an additional noise control strategy is
needed. This is the case, e.g., for the image shown in fig.
\ref{fig:fuoco} (left): the dark area at the bottom right is not
completely uniform, but it is filled with noisy pixels that are
difficult to distinguish. However, the contrast enhancement
produced by our algorithms reveals them, as the unpleasant image in
fig. \ref{fig:fuoco} (center) shows.

To avoid this undesirable effect, we have devised the following
strategy. First of all, we apply on the input image an \emph{extrema killer}, or Luc Vincent grain filter, to remove intensity peaks and valleys, see \cite{Soille:99}. Secondly, we apply one of our
color enhancement algorithms to the pre-filtered images. Finally, we add back
the details that were removed in the first step.

Our tests have shown that this procedure works in a satisfactory way for
very dark images as the one in fig. \ref{fig:fuoco}, but it tends
to slightly blur brighter images or pictures with a greater amount
of details. For this reason, a user can choose to add or not this
strategy to the filtering procedure, depending on the particular
need or on image features. Possibly, an automatization of this choice could
also be possible: the user could define a threshold limit for global
image intensity and variance, in order to filter only those
pictures whose intensity and amount of detail (measured, e.g., by
variance) are below a certain level.

\section{Conclusions and future works}

We have presented an investigation about perceptually inspired color enhancement from the point of view of variational techniques. Instead of explicitly devising an algorithm inspired by some Human Visual System features, we have translated the basic phenomenology of human color vision into mathematical assumptions to be fulfilled by perceptual energy functionals. Operative algorithms for perceptual color correction can then be obtained by minimizing this kind of functionals.

The advantage to analyze perceptual energy functionals is that they can put in evidence in a clearer way the action of color correction in terms of important image features such as, e.g., contrast and dispersion. In fact, our set of assumptions determines a particular class of functionals, whose minimization corresponds to two opponent mechanisms: on one side a local and non-linear contrast enhancement, on the other side a control of dispersion around the middle gray and around the input image data values. The minimization is implemented through an iterative gradient descent technique; the stability of this method and the existence of minima have been proven.

We also pointed out theoretical similarities and differences between these models and existing perceptual color correction models, Retinex and ACE, by identifying, in the class of perceptual functionals, an energy functional that leads to ACE and one that corresponds to a sort of symmetrization of Retinex.

Tests on different images confirm theoretical predictions about the proposed algorithms, which can enhance details and remove color cast without introducing noise. However, already present noise can eventually be amplified. For this reason we also provided a simple and fast pre and post processing procedure that can be integrated in the algorithms to avoid noise amplification in very dark images.

The algorithms depend on four explicit parameters with a clear meaning. Even though there is a set of parameters that works in a satisfactory way for all images of our test set, an image dependent tuning could possibly improve the enhancement characteristics. However, this tuning is beyond the scope of this paper and we consider it as an open problem.

Finally, we believe that the noise control procedure presented in this paper can be improved, e.g. treating the problem from a functional point of view and investigating the relationship between noise and local spatial frequency.

\section*{Acknowledgments}

R. Palma-Amestoy acknowledges support by Alfa CVFA, AML//  19.0902/97/0666/II-0366-FA. E. Provenzi acknowledges partial
support by PRIN-MIUR research project, 2005115173-002. M. Bertalm\'\i o and V. Caselles acknowledge partial support by PNPGC project, reference MTM2006-14836, and IP-RACINE Project IST-511316.

\section*{Appendix}\label{proof:var_minmax}

\subsection{The proof of  Proposition \ref{prop:existence}}\label{proof:var_minmax0}

Notice that since $c:(0,1]\times (0,1] \to \mathbb{R}$ is continuous, then
$c(I(x),I(y))$ is a bounded function for any $I \in {\cal F}_\rho : = \{I:{\mathfrak{I}} \to [0,1], \, I(x) \geq \rho \; \forall x\in \igot\}$.
Then the term $C_w(I)$ is bounded in ${\cal F}_\rho$. The same can be said of $D_{\alpha,\beta}^{\cal E}(I)$.
In particular, $\inf_{I\in {\cal F}_\rho} E(I) > -\infty$. Let $I_k\in {\cal F}_\rho$ be a minimizing sequence, i.e., assume that
$E(I_k) \to  \inf_{I\in {\cal F}_\rho} E(I)$ as $k\to\infty$. Since $I_k$ is uniformly bounded and we are
working in a finite dimensional space, there exists a subsequence, denote it again by $I_k$, that converges to some
$I_\infty\in {\cal F}_\rho$.
Since both terms $D_{\alpha,\beta}^{\cal E}(I)$ and $C_w(I)$ are continuous in ${\cal F}_\rho$, then
$E(I_\infty) = \inf_{I\in {\cal F}_\rho} E(I)$. The corresponding result for the quadratic dispersion
term $D_{\alpha,\beta}^q(I)$ is proved in the same way.

\subsection{Examples of nice regularization of the absolute value}\label{proof:var_minmax1}

Let us first check that the examples of regularization of $A(z)$ given in subsection \ref{subsect:minEf}
are nice regularization. Recall that we denote by $O(F(\epsilon))$ any expression satisfying $\vert O(F(\epsilon))\vert \leq C F(\epsilon)$
for some $C > 0$ and $\epsilon > 0$ small enough. The examples are:

\medskip\noindent a) $A_\epsilon(z) = \sqrt{\epsilon^2+\vert z\vert^2}-\epsilon$:
in this case $s_\epsilon(z) = \frac{z}{\sqrt{\epsilon^2+\vert z\vert^2}}$.
Then $\vert A_\epsilon(z) - \vert z\vert \vert = \epsilon + \vert z\vert - \sqrt{\epsilon^2+\vert z\vert^2}\leq \epsilon$, that is
$Q_{1,\epsilon}(z) = O(\epsilon)$. On the other hand,
$Q_{2,\epsilon}(z) := A_\epsilon(z) - z s_\epsilon(z) = \frac{\epsilon^2}{\sqrt{\epsilon^2 + z^2}} - \epsilon = O(\epsilon)$
as $\epsilon \to 0$ uniformly in $z\in [-1,1]$. The other conditions are immediate and we do not give the details.

\medskip\noindent
b)  $A_\epsilon(z) = z \frac{\arctan(z/\epsilon)}{\arctan (1/\epsilon)} -
\frac{\epsilon}{2\arctan (1/\epsilon)} {\rm log} (1+\frac{z^2}{\epsilon^2})$: in this case
$s_\epsilon(z) = \frac{\arctan(z/\epsilon)}{\arctan (1/\epsilon)}$. Then
$$A_\epsilon(z) = \vert z\vert + \vert z\vert \frac{(\arctan(\vert z\vert /\epsilon) - \arctan(1/\epsilon))}{\arctan(1/\epsilon)}
- \frac{\epsilon/2{\rm log} (1+\frac{z^2}{\epsilon^2})}{\arctan(1/\epsilon)}.
$$
Since we have for any $z\in [-1,1]$,
$$
\left \vert \arctan(1/\epsilon) - \arctan(\vert z\vert/\epsilon) \right\vert = \int_{\vert z\vert/\epsilon}^{1/\epsilon}
\frac{1}{1+s^2} \, ds \leq \int_{\vert z\vert/\epsilon}^{1/\epsilon}
\frac{1}{s^2} \, ds = \epsilon\left( \frac{1}{\vert z\vert}-1 \right),
$$
we have
$$
Q_{1,\epsilon}(z) = \vert A_\epsilon(z) - \vert z\vert \vert \leq \frac{\epsilon}{\arctan(1/\epsilon)} + \frac{\epsilon/2{\rm log} (1+\frac{z^2}{\epsilon^2})}{\arctan(1/\epsilon)} = O(\epsilon \log \,  (1/\epsilon))
$$
uniformly in $z\in [-1,1]$. Finally,
$$Q_{2,\epsilon}(z) = A_\epsilon(z) - z s_\epsilon(z) = - \frac{\epsilon/2{\rm log} (1+\frac{z^2}{\epsilon^2})}{\arctan(1/\epsilon)}  =
O(\epsilon \log \,  (1/\epsilon)),$$
uniformly in $z\in [-1,1]$. Again, the other conditions are immediate and we do not give the details.

\subsection{The proof of  Proposition \ref{prop:var_minmax}}\label{proof:var_minmax2}

Before going into the proof of Proposition \ref{prop:var_minmax}, in the following Lemma
we compute the first variation of a general functional in presence of symmetries.

\medskip

\begin{lemma}\label{lemma:symm}
	Let $w:\mathfrak{I}^2\to\mathbb{R}$ be a symmetric function in $(x,y)$ and $F:(0,\infty)^2\to\mathbb{R}$ be a
	differentiable function in its variables $(a,b)$. Assume that $F$ is symmetric, i.e.
	$F(a,b) = F(b,a)$ for any $(a,b) \in (0,\infty)^2$.
	Let $F_1(a,b) = \frac{\partial F}{\partial a}(a,b)$. Then,
	given \beq \label{def:deneralE} E(I)=\iint_{\mathfrak I^2} w(x,y) F(I(x),I(y)) dx
	dy,\eeq its first variation can be written as
	\begin{equation}\label{eq:symm}
		\delta E(I) = 2 \int_{\mathfrak I} w(x,y) F_1(I(x),I(y))\,  dy.
	\end{equation}
\end{lemma}

\medskip

\proof Let $F_2(a,b) = \frac{\partial F}{\partial b}(a,b)$. Since $F(a,b) = F(b,a)$, for all
$a,b > 0$, we have
\beq \label{sym12}
F_1(a,b) = F_2(b,a).
\eeq
By definition, the first variation of $E(I)$ in the direction $\delta I$ is
\beq \delta E(I,\delta I) =
\iint_{\mathfrak I^2} \, w(x,y) F_1(I(x),I(y)) \, \delta I(x) \,
dx dy + \iint_{\mathfrak I^2} w(x,y) \, F_2(I(x),I(y)) \, \delta
I(y) \, dx dy.
\eeq
Interchanging the role of $x$ and $y$ in the
second integral of the equation above and using (\ref{sym12}) we get
\beq
\iint_{\mathfrak I^2} w(x,y) \, F_2(I(y),I(x)) \, \delta I(x) \,
dx dy = \iint_{\mathfrak I^2} w(x,y) \, F_1(I(x),I(y)) \, \delta
I(x) \, dx dy
\eeq
so that
\beq \delta E(I,\delta I) =
\int_{\mathfrak I} \left( 2 \int_{\mathfrak I} w(x,y) \,
F_1(I(x),I(y))  \right) \delta I(x)
\, dx
\eeq
and the proposition follows. \qed

\noindent {\em Proof of Proposition \ref{prop:var_minmax}.}
We assume that $A_\epsilon(z)$ is a nice regularization of the absolute value
$A(z) = \vert z\vert$, $z\in \mathbb{R}$. We define the regularization
${\rm max}_\epsilon(a,b) =  \frac{1}{2} (a+b-A_\epsilon(a,b))$ and ${\rm min}_\epsilon(a,b)
=  \frac{1}{2} (a+b+ A_\epsilon(a,b))$ for any $a,b\in\mathbb{R}$. Since the integrand in
$C^{\, {\rm id}}_{w,\epsilon}(I), C^{\, {\rm log}}_{w,\epsilon}(I), C^{\, -{\cal M}}_{w,\epsilon}(I)$
is differentiable, the result follows by applying Lemma\ref{lemma:symm}.

\noindent $(i)$ The functional $C^{\, {\rm id}}_{w,\epsilon}(I)$ can be written as
(\ref{def:deneralE}) taking
$F(a,b)= \frac{1}{4} \frac{{\rm min}_\epsilon(a,b)}{{\rm max}_\epsilon(a,b)}$.
Since $$F_1(a,b) = \frac{1}{8} \frac{A_\epsilon(a-b)-s_\epsilon(a-b) (a+b)}{{\rm max}_\epsilon(a,b)^2}
= - \frac{1}{4} \frac{b}{{\rm max}_\epsilon(a,b)^2} s_\epsilon(a-b) + \frac{1}{8} \frac{Q_{2,\epsilon}(a-b)}{{\rm max}_\epsilon(a,b)^2},$$
by Lemma \ref{lemma:symm}, we have
\beq
\begin{array}{ll}
	\delta C^{\,
		\rm id}_{w,\epsilon}(I) = - \frac{1}{2}\sum_{y\in \mathfrak I} w(x,y) \left(\frac{I(y)s_\epsilon(I(x)-I(y))}{{\rm max}_\epsilon(I(x),I(y))^2}
	- \frac{1}{2} \frac{Q_{2,\epsilon}(I(x)-I(y))}{{\rm max}_\epsilon(I(x),I(y))^2}\right)\\
	\\
	\qquad \qquad = - \frac{1}{2}\sum_{y\in \mathfrak I} w(x,y) \frac{I(y)}{{\rm max}_\epsilon(I(x),I(y))^2}s_\epsilon(I(x)-I(y))
	+ S_\epsilon,
\end{array}
\eeq
where $S_\epsilon = O(Q_{2,\epsilon}(I(x)-I(y)))$, Now, since $A_\epsilon(z) = \vert z\vert + Q_{1,\epsilon}(z)$, then
${\rm max}_\epsilon(a,b) = {\rm max}(a,b) + \frac{1}{2} Q_{1,\epsilon}(z)$, and we may write
$$
\frac{1}{{\rm max}_\epsilon(a,b)^2} = \frac{1}{{\rm max}(a,b)^2} + O(Q_{1,\epsilon}(a,b))
$$
where the expression $Q_{1,\epsilon}(a,b)$ is uniform in $a,b \in [\rho,1]$. Thus, we obtain
\beq \delta C^{\,
	\rm id}_{w,\epsilon}(I) = - \frac{1}{2}\sum_{y\in \mathfrak I} w(x,y) \frac{I(y)}{{\rm max}(I(x),I(y))^2}s_\epsilon(I(x)-I(y))
+ S'_\epsilon,
\eeq
where $S'_\epsilon= O(Q_{1,\epsilon}(I(x)-I(y))+ Q_{2,\epsilon}(I(x)-I(y)))$.

\noindent $(ii)$ The functional $C^{\, {\rm log}}_{w,\epsilon}(I)$ can be written as
(\ref{def:deneralE}) taking $F(a,b)= \frac{1}{4} \log \frac{{\rm min}_\epsilon(a,b)}{{\rm max}_\epsilon(a,b)}$. Since
\begin{eqnarray*}
	F_1(a,b) & = &  \frac{1}{8}  \frac{A_\epsilon(a-b)-s_\epsilon(a-b) (a+b)}{{\rm min}_\epsilon(a,b){\rm max}_\epsilon(a,b)}
	\\ & = & - \frac{1}{4} \frac{b}{{\rm min}_\epsilon(a,b){\rm max}_\epsilon(a,b)} s_\epsilon(a-b) + \frac{1}{8} \frac{Q_{2,\epsilon}(a-b)}{{\rm min}_\epsilon(a,b){\rm max}_\epsilon(a,b)},
\end{eqnarray*}
by Lemma \ref{lemma:symm}, we have
\beq \delta C^{\,
	\log}_{w,\epsilon}(I) = - \frac{1}{2}\sum_{y\in \mathfrak I} w(x,y) \left(\frac{I(y)s_\epsilon(I(x)-I(y))}{(I(x)\cdot I(y))_\epsilon}
- \frac{1}{2} \frac{Q_{2,\epsilon}(I(x)-I(y))}{(I(x)\cdot I(y))_\epsilon}\right),
\eeq
where $(a\cdot b)_\epsilon : = {\rm min}_\epsilon(a,b){\rm max}_\epsilon(a,b)$, $a,b\in \mathbb{R}$.
Now, since $A_\epsilon(z) = \vert z\vert + Q_{1,\epsilon}(z)$, we have
$$
\frac{1}{{\rm min}_\epsilon(a,b){\rm max}_\epsilon(a,b)} = \frac{1}{{\rm min}(a,b){\rm max}(a,b)} + O(Q_{1,\epsilon}(a,b)) =
\frac{1}{ab} + O(Q_{1,\epsilon}(a,b))
$$
where the expression $Q_{1,\epsilon}(a,b)$ is uniform in $a,b \in [\rho,1]$.
Thus, we obtain
\beq
\delta C^{\,
	\log}_{w,\epsilon}(I) = - \frac{1}{2}\sum_{y\in \mathfrak I} w(x,y) \frac{1}{I(x)}s_\epsilon(I(x)-I(y))
+ S_\epsilon,
\eeq
where $S_\epsilon= O(Q_{1,\epsilon}(I(x)-I(y))+ Q_{2,\epsilon}(I(x)-I(y)))$.

\noindent $(iii)$ The functional $C^{\, -{\cal M}}_{w,\epsilon}(I)$ can be written as
(\ref{def:deneralE}) taking $F(a,b)= - \frac{1}{4}\frac{A_\epsilon(a - b)}{a+b}$. Since
$$F_1(a,b) = - \frac{1}{2} \frac{b}{(a+b)^2} s_\epsilon(a-b)+ \frac{1}{4}
\frac{Q_{2,\epsilon}(a-b)}{(a+b)^2},$$
by Lemma \ref{lemma:symm}, we have
\begin{eqnarray*}
	\delta C^{-{\cal M}}_{w,\epsilon}(I) & = & - \sum_{y\in \mathfrak I} w(x,y) \left(\frac{I(y)s_\epsilon(I(x)-I(y))}{(I(x)+ I(y))^2}
	- \frac{1}{2}\frac{Q_\epsilon(I(x)-I(y))}{(I(x)+ I(y))^2}\right) \\
	& = & - \sum_{y\in \mathfrak I} w(x,y) \frac{I(y)}{(I(x)+ I(y))^2} s_\epsilon(I(x)-I(y)) + S_\epsilon,
\end{eqnarray*}
where $S_\epsilon= O(Q_{2,\epsilon}(I(x)-I(y)))$.
\medskip

\subsection{Remarks on the stability of the algorithm}\label{sec:stability}

The general form of semi-implicit gradient descent equation is
\beq\label{eq:generalform2} I^{k+1}(x) = \frac{I^k(x) + \Delta t \left( \frac{\alpha}{2} +
	\beta I_0(x) + \frac{1}{2} R_{I^k}(x)\right)}{1+\Delta t(\alpha + \beta)}, \eeq
being $R_{I}(x)=\sum_{y\in {\mathfrak I}} w(x,y) \, r(I(x),I(y))$, for a suitable integrand function $r$. To fix ideas, we
consider the case where
\beq \label{eq:ridstability}
R^{\, \rm id}_{\epsilon,I}(x) := \sum_{y\in \mathfrak I} w(x,y)
\frac{{\rm min}(I(x),I(y))}{{\rm max}(I(x),I(y))}s_\epsilon(I(x)-I(y)).
\eeq

We assume that $I_0:\mathfrak{I}\to [\rho,1]$ where $\rho > 0$ is a minimum value for the initial image.
For us $\rho = 1/255$, and this means that we assume that $I_0$ does not take the value $0$.
Our purpose is to prove the following statements.

\medskip

\begin{lemma}\label{lemma:stability} Assume that $\alpha \geq \frac{1}{1-2\rho} > 0$. Then
	
	\noindent $(i)$ $ \rho \leq I^k(x) \leq 1$ for any $x\in\mathfrak{I}$ and any $k\geq 1$.
	
	\noindent $(ii)$ $\Vert I^{k+1}- I^k\Vert_p \leq \frac{1 + \Delta t \left( \frac{1}{\rho} + m_\epsilon \right)}{1 + \Delta t (\alpha+\beta)} \Vert I^{k}- I^{k-1}\Vert_p$ for any $p\in [1,\infty]$ and any $k\geq 1$,
	where $m_\epsilon := \max_{z\in [-1,1]} \vert s_\epsilon^\prime(z)\vert$. Thus, if $\alpha + \beta > \frac{1}{\rho} + m_\epsilon $, then
	the iterative scheme given in (\ref{eq:generalform2}) is convergent to the unique function $I^*$ satisfying
	\beq\label{AELE}
	\alpha (I^*-\frac{1}{2}) + \beta(I^*-I_0) - \frac{1}{2} R^{\rm id}_{\epsilon,I^*} = 0.
	\eeq
\end{lemma}

\medskip
As usual, the $\Vert \cdot\Vert_p$ norm of a vector $v = (v_i)_{i=1}^n \in \mathbb{R}^n$, $n\geq 1$, is defined by
$
\Vert v\Vert_p : = \left(\sum_{i=1}^n \vert v_i\vert^p\right)^{1/p}$ if $p\in [1,\infty)$,
and
$ \Vert v\Vert_\infty := {\rm max}_{i=1,\ldots,n} \vert v_i\vert$ if $p=\infty$.

\medskip Notice that (\ref{AELE}) is essentially (and not exactly, due
to our regularization of the basic contrast variable) the Euler-Lagrange equation corresponding to the energy $I^{\rm id}_{w,\epsilon}$.
Notice also that $\alpha + \beta > \frac{1}{\rho} + m_\epsilon $ is not a reasonable condition in practice. The reason is that
$1/\rho = 255$ and $m_\epsilon\approx 1/\epsilon$ for the particular nice regularization of $A(z)$ given in
the examples a),b) given in subsection \ref{subsect:minEf}. Then the values of $\alpha$ and/or $\beta$
are too big and produce a strong attachment to the initial data and/or the value $1/2$; in this case we do not have enough
enhancement power. The values of $\alpha$ and  $\beta$ used in practice are much smaller, $\alpha = 255/253$ (in accordance
with the assumption of Lemma \ref{lemma:stability}) and $\beta = 1$, and the algorithm exhibited stability and convergence in practice.

\medskip To prove Lemma \ref{lemma:stability}, we need the auxiliary result.

\medskip

\begin{lemma}\label{ausiliar:stab}
	Let $I, \tilde{I}: \mathfrak{I}\to [\rho,1]$. Then
	
	\noindent
	$(i)$ $-1\leq R^{\, \rm id}_{\epsilon,I}(x)\leq 1$ for any $x\in \mathfrak{I}$.
	
	\noindent
	$(ii)$ $\Vert R^{\, \rm id}_{\epsilon,I} - R^{\, \rm id}_{\epsilon,\tilde{I}} \Vert_p
	\leq 2\left( \frac{1}{\rho} + m_\epsilon \right)
	\Vert I - \tilde{I}\Vert_p$
	
\end{lemma}

\medskip\noindent {\em Proof.} $(i)$ The first statement follows easily from
\beq \label{eq:ridstability1}
\vert R^{\, \rm id}_{\epsilon,I}(x) \vert \leq  \sum_{y\in \mathfrak I} w(x,y)
\frac{{\rm min}(I(x),I(y))}{{\rm max}(I(x),I(y))} \,  \vert s_\epsilon(I(x)-I(y))\vert \leq \sum_{y\in \mathfrak I} w(x,y) = 1 .
\eeq

\smallskip\noindent $(ii)$ We have
\begin{equation}\label{p:est0}
	\vert R^{\, \rm id}_{\epsilon,I}(x) - R^{\, \rm id}_{\epsilon,\tilde{I}}(x) \vert \leq
	\sum_{y\in \mathfrak I} w(x,y)
	\left \vert \frac{{\rm min}(I(x),I(y))}{{\rm max}(I(x),I(y))} \,   s_\epsilon(I(x)-I(y))
	- \frac{{\rm min}(\tilde I(x),\tilde I(y))}{{\rm max}(\tilde I(x),\tilde I(y))} \,   s_\epsilon(\tilde I(x)-\tilde I(y))  \right \vert.
\end{equation}
Let us prove that for any $a,b,c,d, \in [\rho,1]$ we have
\begin{equation}\label{p:est1}
	\left \vert \frac{{\rm min}(a,b)}{{\rm max}(a,b)} \,   s_\epsilon(a-b))
	- \frac{{\rm min}(c,d)}{{\rm max}(c,d)} \,   s_\epsilon(c-d)  \right \vert  \leq \left(\frac{1}{\rho} +  m_\epsilon\right)
	(\vert a-c \vert + \vert b-d \vert),
\end{equation}
where $m_\epsilon  = \max_{z\in [-1,1]} \vert s_\epsilon^\prime(z)\vert$.
Indeed, we have
\begin{equation}\label{p:est2}
	\begin{array}{ll}
		\left \vert \frac{{\rm min}(a,b)}{{\rm max}(a,b)} \,   s_\epsilon(a-b))
		- \frac{{\rm min}(c,d)}{{\rm max}(c,d)} \,   s_\epsilon(c-d)  \right \vert   \leq
		\left \vert \frac{{\rm min}(a,b)}{{\rm max}(a,b)}
		- \frac{{\rm min}(c,d)}{{\rm max}(c,d)}\right \vert  \left \vert    s_\epsilon(c-d)  \right \vert \\ \\
		\qquad \qquad \qquad \qquad \qquad \qquad
		\qquad \qquad  +  \left\vert \frac{{\rm min}(c,d)}{{\rm max}(c,d)} \right\vert \left \vert s_\epsilon(a-b)  - s_\epsilon(c-d) \right\vert.
	\end{array}
\end{equation}
Now, we apply the intermediate value theorem to the function $F_\epsilon (z_1,z_2) =
\frac{{\rm min}_\epsilon(z_1,z_2)}{{\rm max}_\epsilon(z_1,z_2)}$,
$z_1,z_2\in [\rho,1]$ and we obtain
$$
\vert F_\epsilon (a,b) - F_\epsilon (c,d) \vert \leq \max_{i=1,2} \max_{z_1,z_2\in [\rho,1]}
\left \vert \frac{\partial}{\partial z_i}  F_\epsilon (z_1,z_2)\right \vert
(\vert a-c \vert + \vert b-d \vert)
$$
Letting $\epsilon\to 0$ we obtain,
\begin{equation}\label{p:est3}
	\left \vert \frac{{\rm min}(a,b)}{{\rm max}(a,b)} - \frac{{\rm min}(c,d)}{{\rm max}(c,d)}\right\vert
	\leq \frac{1}{\rho} (\vert a-c \vert + \vert b-d \vert).
\end{equation}
Introducing (\ref{p:est3}) in (\ref{p:est2}) and using again the intermediate value theorem with $s_\epsilon$ we
obtain (\ref{p:est1}). Now, introducing (\ref{p:est1}) into (\ref{p:est0}) we obtain
\begin{equation}\label{p:est4}
	\vert R^{\, \rm id}_{\epsilon,I}(x) - R^{\, \rm id}_{\epsilon,\tilde{I}}(x) \vert \leq \left(\frac{1}{\rho} +  m_\epsilon\right)
	\left ( \vert I(x)-\tilde{I}(x)\vert + \sum_{y\in \mathfrak I} w(x,y) \vert I(y)-\tilde{I}(y)\vert  \right)
\end{equation}
Since the function $r\in [0,\infty) \to r^p$, $p\geq 1$, is convex, using Jensen's inequality we have
\begin{eqnarray*}
	\left( \sum_{x\in\mathfrak{I}} \left(\sum_{y\in \mathfrak I} w(x,y) \vert I(y)-\tilde{I}(y)\vert \right)^p \,  \right)^{1/p}
	& \leq & \left( \sum_{x\in\mathfrak{I}} \sum_{y\in \mathfrak I} w(x,y) \vert I(y)-\tilde{I}(y)\vert^p \,  \right)^{1/p} \\
	& \leq & \left( \sum_{y\in \mathfrak I}\vert I(y)-\tilde{I}(y)\vert^p \,  \right)^{1/p}.
\end{eqnarray*}
Using this inequality in (\ref{p:est4}) we obtain
$$
\Vert R^{\, \rm id}_{\epsilon,I} - R^{\, \rm id}_{\epsilon,\tilde{I}} \Vert_p \leq
2 \left(\frac{1}{\rho} +  m_\epsilon\right)   \Vert I-\tilde{I}\Vert_p
$$
for any $p\in [1,\infty)$. Letting $p\to\infty$ we also obtain $(ii)$ for $p=\infty$.

\medskip\noindent {\em Proof of Lemma \ref{lemma:stability}.} $(i)$
First, let us check that all iterations $I^k(x) \leq 1$ for any $x\in\mathfrak{I}$ and any $k\geq 1$.
This follows easily by induction from (\ref{eq:generalform2}) since, assuming that $I^k(x) \leq 1$ and
using that $\alpha \geq 1$ and Lemma (\ref{ausiliar:stab}), $(i)$, we have
$$
I^{k+1}(x) \leq  \frac{1 + \Delta t \left( \frac{\alpha}{2} +
	\beta  + \frac{1}{2} \right)}{1+\Delta t(\alpha + \beta)} \leq
\frac{1+\Delta t(\alpha + \beta)}{1+\Delta t(\alpha + \beta)} = 1.
$$
Now assume that $I^k(x)\geq \rho$. To prove that $I^{k+1}(x)\geq \rho$, we use that $R^{\, \rm id}_{\epsilon,I}(x)\geq -1$ in
(\ref{eq:generalform2}) and we obtain,
$$
I^{k+1}(x) \geq  \frac{\rho + \Delta t \left( \frac{\alpha}{2} +
	\beta \rho - \frac{1}{2} \right)}{1+\Delta t((\alpha -1)/2+ \beta)} \geq
\rho \frac{1+\Delta t(\alpha + \beta)}{1+\Delta t(\alpha + \beta)} = \rho,
$$
because we are assuming that $\frac{\alpha}{2}-\frac{1}{2} \geq \alpha\rho$.

\medskip\noindent $(ii)$ We write
\beq\label{eq:generalform3}
I^{k}(x) = \frac{I^{k-1}(x) + \Delta t \left( \frac{\alpha}{2} +
	\beta I_0(x) + \frac{1}{2} R^{\rm id}_{\epsilon,I^{k-1}}(x)\right)}{1+\Delta t(\alpha + \beta)},
\eeq
and take the difference of (\ref{eq:generalform2}) and (\ref{eq:generalform3}), to obtain
\begin{equation*}
	\Vert I^{k+1} - I^k \Vert_p \leq \frac{\Vert I^{k} - I^{k-1} \Vert_p}{1 + \Delta t (\alpha+\beta)}
	+ \frac{1}{1 + \Delta t (\alpha+\beta)} \frac{1}{2} \Vert R^{\rm id}_{\epsilon,I^{k}} - R^{\rm id}_{\epsilon,I^{k-1}}\Vert_p
\end{equation*}
Using Lemma \ref{ausiliar:stab}, $(ii)$, we obtain
$$\Vert I^{k+1}- I^k\Vert_p \leq \frac{1 + \Delta t \left( \frac{1}{\rho} + m_\epsilon \right)}{1 + \Delta t (\alpha+\beta)} \Vert I^{k}- I^{k-1}\Vert_p$$
for any $p\in [1,\infty]$ and any $k\geq 1$. The above estimate is contractive when
$\frac{1 + \Delta t \left( \frac{1}{\rho} + m_\epsilon \right)}{1 + \Delta t (\alpha+\beta)} < 1$, i.e, when
$\alpha + \beta > \frac{1}{\rho} + m_\epsilon$. In this case, the sequence $I^k$ converges to a function $I^*$ satisfying
\begin{equation*}
	I^*(x) = \frac{I^*(x) + \Delta t \left( \frac{\alpha}{2} +
		\beta I_0(x) + \frac{1}{2} R^{\rm id}_{\epsilon,I^*}(x)\right)}{1+\Delta t(\alpha + \beta)},
\end{equation*}
that is, (\ref{AELE}) holds.

\medskip To prove the uniqueness of solutions of (\ref{AELE}), assume that
$I^*$ and $\tilde{I}^*$ are two solutions. Then, we have
$$
I^*- \tilde{I}^* = \frac{1}{\alpha+\beta} \frac{1}{2} (R^{\rm id}_{\epsilon,I^*}-R^{\rm id}_{\epsilon,\tilde{I}^*}).
$$
Hence, taking norms and using Lemma \ref{ausiliar:stab}, $(ii)$, we have
$$
\Vert I^*- \tilde{I}^* \Vert \leq \frac{\frac{1}{\rho}+m_\epsilon}{\alpha+\beta} \Vert I^*- \tilde{I}^* \Vert.
$$
Since $\alpha + \beta > \frac{1}{\rho}+m_\epsilon$ this is a contradiction, unless $I^* = \tilde{I}^*$.


\end{document}